\documentclass[10pt]{article} 
\usepackage[preprint]{rlc}

\usepackage{amssymb}            
\usepackage{mathtools}          
\usepackage{mathrsfs}           
\mathtoolsset{showonlyrefs}     
\usepackage{graphicx}           
\usepackage{subcaption}         
\usepackage[space]{grffile}     
\usepackage{url}                
\usepackage{xspace}
\usepackage{amsmath,amsfonts,bm,amssymb,dsfont}
\usepackage[export]{adjustbox}
\usepackage{wrapfig}

\newcommand{\method}{SplAgger\xspace}
\newcommand{\E}{\mathbb{E}}
\DeclareMathOperator*{\argmax}{arg\,max}

\title{\method: Split Aggregation for\\Meta-Reinforcement Learning}

\author{Jacob Beck  \\
    jacob\_beck@alumni.brown.edu \\
    Department of Computer Science \\
    University of Oxford
    \And
    Matthew Jackson \\
    jackson@robots.ox.ac.uk \\
    Department of Engineering Science \\
    University of Oxford
    \And
    Risto Vuorio \\
    risto.vuorio@keble.ox.ac.uk \\
    Department of Computer Science \\
    University of Oxford
    \And
    Zheng Xiong \\
    zheng.xiong@cs.ox.ac.uk \\
    Department of Computer Science \hspace{.15cm} \\
    University of Oxford
    \And
    Shimon Whiteson \\
    shimon.whiteson@cs.ox.ac.uk \\
    Department of Computer Science \\
    University of Oxford}

\begin{document}

\maketitle

\begin{abstract}
A core ambition of reinforcement learning (RL) is the creation of agents capable of rapid learning in novel tasks.
Meta-RL aims to achieve this by directly learning such agents.
\textit{Black box methods} do so by training off-the-shelf sequence models end-to-end.
By contrast, \textit{task inference methods} explicitly infer a posterior distribution over the unknown task, typically using distinct objectives and sequence models designed to enable task inference. 
Recent work has shown that task inference methods are not necessary for strong performance.
However, it remains unclear whether task inference sequence models are beneficial even when task inference objectives are not.
In this paper, we present evidence that task inference sequence models are indeed still beneficial.
In particular, we investigate sequence models with \emph{permutation invariant aggregation}, which exploit the fact that, due to the Markov property, the task posterior does not depend on the order of data.
We empirically confirm the advantage of permutation invariant sequence models without the use of task inference objectives.
However, we also find, surprisingly, that there are multiple conditions under which permutation variance remains useful.
Therefore, we propose \method, which uses both permutation variant and invariant components to achieve the best of both worlds, outperforming all baselines evaluated on continuous control and memory environments. Code is provided at \url{https://github.com/jacooba/hyper}.
\end{abstract}

\section{Introduction}

A prevalent method for creating agents that can quickly learn involves teaching them how to learn quickly.
This problem is well studied under the name of meta-reinforcement learning \citep{beck2023survey}.
In meta-RL, an agent learns a reinforcement learning algorithm over a distribution of reinforcement learning problems, called tasks.
In order to condition on data from new tasks, the agent can use a generic sequence model, such as a recurrent neural network (RNN), trained end-to-end \citep{duan2016rl,wang2016learning}.
These methods are referred to as \textit{black-box} methods \citep{beck2023survey}.

By contrast, a distinct category of research focuses on methods specialized for meta-RL.
These methods typically infer an explicit posterior over tasks, given data collected from a new task.
To do so, they generally use distinct objectives and distinct sequence models, designed to enable inference of the unknown task.
In particular, it is common to use sequence models that are invariant to the order of their inputs, which we refer to as permutation invariant aggregation.
Due to the Markov property, the true posterior over tasks does not depend on this order.
Collectively, these methods are referred to as \textit{task-inference} methods \citep{beck2023survey}.

While many task inference methods have been developed for meta-RL, recent work has shown black-box methods to be more effective in practice \citep{ni2022recurrent,beck2023recurrent}.
However, these results focus primarily on demonstrating the superiority of the end-to-end objective used in black-box methods over the task-inference objective, and do not investigate the effect of the particular sequence model.
This leaves an open question: \textit{When using an end-to-end objective, is it still worth using permutation invariant sequence models?}

In this paper, we answer in the affirmative.
We show that permutation invariant sequence models still confer an advantage in a number of domains, even when trained end-to-end.
However, we also find, surprisingly, that there are domains where dependence on the permutation remains useful. 
Specifically, we find sequence models with a permutation variant component to be less sensitive to choices in the permutation invariant component, and we find permutation variance useful when there exist permutation variant suboptimal policies. 
We extensively investigate the conditions under which each type of sequence model is useful, conduct analysis to support our conclusions, and propose a simple sequence model, called Split Aggregator, or \method, adapted and simplified from the literature on partial observability \citep{beck2020AMRL}.
\method, depicted in Figure \ref{fig:splag}, uses both permutation invariant and permutation variant components to achieve the best of both worlds and high returns in all domains evaluated.

\section{Related Work}

\paragraph{End-to-End Meta-RL} 
The problem setting defined by meta-RL can be viewed as a particular type of partially observable Markov decision process (POMDP) \citep{beck2023survey}.
From the theory of POMDPs, we know that the optimal policy for POMDPs, and thus meta-RL, can be represented as an arbitrary function of history \citep{subramanian2022approximate}.
Inspired by this, one category of meta-RL methods, called \textit{black-box} methods, train general purpose sequence models end-to-end on the meta-RL objective \citep{duan2016rl,wang2016learning,ni2022recurrent,team2023human,beck2023recurrent}.
Recently, it has been shown that these methods are a strong baseline in meta-RL \citep{ni2022recurrent}.
Moreover, if hypernetworks \citep{ha2017hypernetworks} are used, these methods have superior performance to task inference methods \citep{beck2023recurrent}.
We build off of these results in our paper, using end-to-end trained hypernetworks, following the methods in \citet{beck2022hyper,beck2023recurrent}.
However, in contrast to these papers, we provide strong evidence that specialised sequence models, still trained end-to-end, can provide a strong advantage.

\paragraph{Sequence Models in Meta-RL}
Task-inference methods explicitly attempt to infer a posterior distribution over the identify of the task \citep{beck2023survey}.
Following directly from the Markov property, it can be shown that this posterior does not depend on the order of the data on which the agent conditions.
While generic sequence models, such as RNNs, may model permutation invariance, they must learn to do so.
In order to incorporate this inductive bias directly, methods generally modify the sequence model to be permutation invariant \citep{rakelly2019efficient, galashov2019meta, raileanu2020fast, wang2022learning, imagawa2022off}.
One popular method, called probabilistic embeddings for actor-critic RL, or \textit{PEARL} \citep{rakelly2019efficient}, incorporates permutation invariance into the probability density function of a stochastic latent variable summarizing history.
Specifically, the density function, modelled as a product over individual transitions in the data, is permutation invariant. 
We compare to this style of aggregation in our experiments.
Another approach uses commutative operators applied across the data \citep{imagawa2022off, wang2022learning, galashov2019meta}.
Generally, these can be viewed as (conditional) Neural Processes (CNP) \citep{garnelo2018neural,garnelo2018conditional}, and so we compare to this style of aggregation as well.
Yet another approach uses attention, self-attention, or transformers \citep{mishra2018a,fortunato2019generalization,nguyen2022transformer}. 
Attention is inherently permutation invariant.
Still, attention is computationally expensive: whereas both commutative aggregation and recurrent networks use $O(1)$ memory and compute per timestep, attention generally requires $O(t^2)$ memory and compute per timestep $t$, for autoregressive inference.
While fast approximations of attention  exist \citep{katharopoulos2020transformers}, the computational requirements are significantly larger.
Additionally, our limited experimentation shows they have difficulty learning on our domains (see Appendix \ref{sec:transform} for details).
Thus, we limit our solutions to constant memory and compute, in line with sequence models designed to quickly handle long contexts \citep{garnelo2018neural,beck2020AMRL}.
Finally, we compare to Aggregated Memory for RL, or \textit{AMRL} \citep{beck2020AMRL}.
While AMRL was originally proposed as a method for POMDPs, it was evaluated in meta-RL problem settings.
Our method proposes a simplification to AMRL that is vital in practice.
Details of AMRL are covered in Section \ref{sec:model}.

\paragraph{In-Context Learning}
The methods we investigate in this paper can be seen as performing \textit{in-context} learning.
Learning that occurs after training and within the activations of a sequence model is called \textit{in-context} learning \citep{brown2020language}.
Black box and task inference methods both perform in-context learning.
In part due to the popularity of large language models \citep{devlin2019bert,brown2020language,chowdhery2022palm}, in-context learning has gained significant traction recently, including in decision-making applications \citep{raparthy2023generalization, lee2024supervised} and reinforcement learning (RL) \citep{kirsch2022general}.
While other meta-RL methods exist that do not use sequence models for in-context learning, such as parameterized policy gradient methods, they generally require more samples to adapt to novel tasks, including those used in our benchmarks \citep{zintgraf2021varibad,beck2023survey}.
In-context learning promises to address the sample inefficiency still impeding progress in reinforcement learning.
Learning to perform in-context RL is the problem studied in meta-RL.

\section{Background}
\label{sec:background}

\subsection{Problem Setting}
We formalize the learning problem as a Markov Decision Processes (MDP) represented by the tuple $(\mathcal{S, A, R, P}, \gamma)$, where $\mathcal{S}$ denotes the state space, $\mathcal{A}$ the action space, $\mathcal{R}$ the reward function, $\mathcal{P}$ the state transition probabilities, and $\gamma$ the discount factor.
During each time-step $t$, an agent finds itself in a state $s_t \in \mathcal{S}$, observes this state, and chooses an action $a_t \in \mathcal{A}$. 
The MDP then transitions to a new state $s_{t+1}$, following the probability distribution $s_{t+1} \sim \mathcal{P}(s_{t+1} | s_t, a_t)\colon \mathcal{S} \times \mathcal{A} \times\mathcal{S} \rightarrow \mathbb{R}_+$, and the agent receives a reward $r_{t} = \mathcal{R}(s_t, a_t)\colon \mathcal{S} \times \mathcal{A} \rightarrow \mathbb{R}$.
The agent acts to maximize the expected future discounted reward, $R(\tau) = \sum_{r_t \in \tau}{\gamma^t r_t}$, where $\tau$ denotes the agent's trajectory throughout an episode in the MDP, and $\gamma \in [0, 1)$ 
is a discount factor.
The agent's decisions are guided by a policy $\pi(a|s): \mathcal{S} \times \mathcal{A} \rightarrow \mathbb{R}_+$, a learned function mapping states to a probability distribution over actions.

Meta-RL algorithms learn an RL algorithm, $f(\tau)$, over a distribution of MDPs, or tasks.
$f(\tau)$ maps from the data, $\tau$, sampled from a single MDP, $\mathcal{M} \sim p(\mathcal{M})$, to policy parameters $\phi$. 
As in a single RL task, $\tau$ is a sequence up to time-step $t$ forming a trajectory of states, actions, rewards, and next states, $\tau \in (\mathcal{S} \times \mathcal{A} \times \mathbb{R} \times \mathcal{S})^{t} $.
We can see $\tau$ as a trajectory of transitions, $\tau_0,\tau_1,...,\tau_t$, where $\tau_t$ is shorthand for the transition, $(s_t,a_t,r_t,s_{t+1})$.
Here, $\tau$ may span multiple episodes within a single MDP, since multiple episodes of interaction may be necessary for learning.
Collectively, we refer to these episodes as a \textit{meta-episode}, and use the same symbol, $\tau$, to refer to it.
The policy, $\pi(a|s;\phi=f_\theta(\tau))$, is parameterized by $\phi$.
$f$ is represented as a sequence model, parameterized by $\theta$, which we refer to as the \textit{meta-parameters}.

The objective in meta-RL is to find meta-parameters $\theta$ that maximize the expected sum of the returns in the meta-episode across a distribution of tasks (MDPs):
\begin{align}\label{eq:objective}
    \argmax_{\theta} \E_{\mathcal{M} \sim p(\mathcal{M})} \bigg[ \E_\tau \bigg[ R(\tau) \bigg| \pi(\cdot;\phi=f_\theta(\tau)), \mathcal{M} \bigg]\bigg].
\end{align}

\subsection{Permutation Invariance}
Task inference methods in meta-RL explicitly infer a posterior over tasks.
In meta-RL, the optimal policy can be computed from both the current state, $s_t$, and this posterior, $P(\mathcal{M}|\tau)$ \citep{beck2023survey}.
Following Bayes's rule and the Markov property, this posterior distribution, for a trajectory of length $T$, can be written,
\begin{align}
    P(\mathcal{M}|\tau) 
    &\propto P(\tau|\mathcal{M})P(\mathcal{M}) & &\text{// Bayes's rule} \\
    &= P(\mathcal{M}) \prod_{t=1}^{t=T}P(a_t,r_t,s_{t+1}|s_t,\mathcal{M}), & &\text{// Markov property}
\end{align}
with the full proof in Appendix \ref{sec:proof}. 

In this expression, the posterior, $P(\mathcal{M}|\tau)$, does not depend on the order of the transitions, $\tau_1,..,\tau_t$.
While it is possible to learn each factor, $P(a_t,r_t,s_{t+1}|s_t,\mathcal{M})$, as a function of each individual transition, $\tau_t$, this form is not particularly amenable to inference, since it requires marginalizing over all MDPs at test time.
Still, it is possible to incorporate the permutation invariant structure into the sequence model directly.

Generally this is done through the use of a permutation invariant binary operator, $\bigoplus$, such as a mean or a sum \citep{garnelo2018neural,garnelo2018conditional,galashov2019meta, imagawa2022off, wang2022learning}, applied over the sequence of inputs.
For a given sequence of encoded inputs, $e_1...,e_{t}$, we write the resulting aggregated representation, $g(e_1,...,e_{t})$:
\begin{equation}
    g(e_1,...,e_{t}) = e_1 \bigoplus e_2  \bigoplus ... \bigoplus e_{t}.
\end{equation}
The benefit of using these operators, in addition to permutation invariance, is that the sequence can be computed recursively, since $g(e_1,...,e_{t}) = g(e_1,...,e_{t-1}) \bigoplus e_t$.
This means that the sequence at all points in time can be compressed into $O(1)$ memory and each new timestep can be computed in $O(1)$ time.
Still, as we will demonstrate in practice, permutation invariance can be beneficial in some environments and detrimental in others.

\begin{figure}[t]
    \centering
    \begin{subfigure}[b]{0.32\textwidth}
        \centering
        \includegraphics[height=4cm]{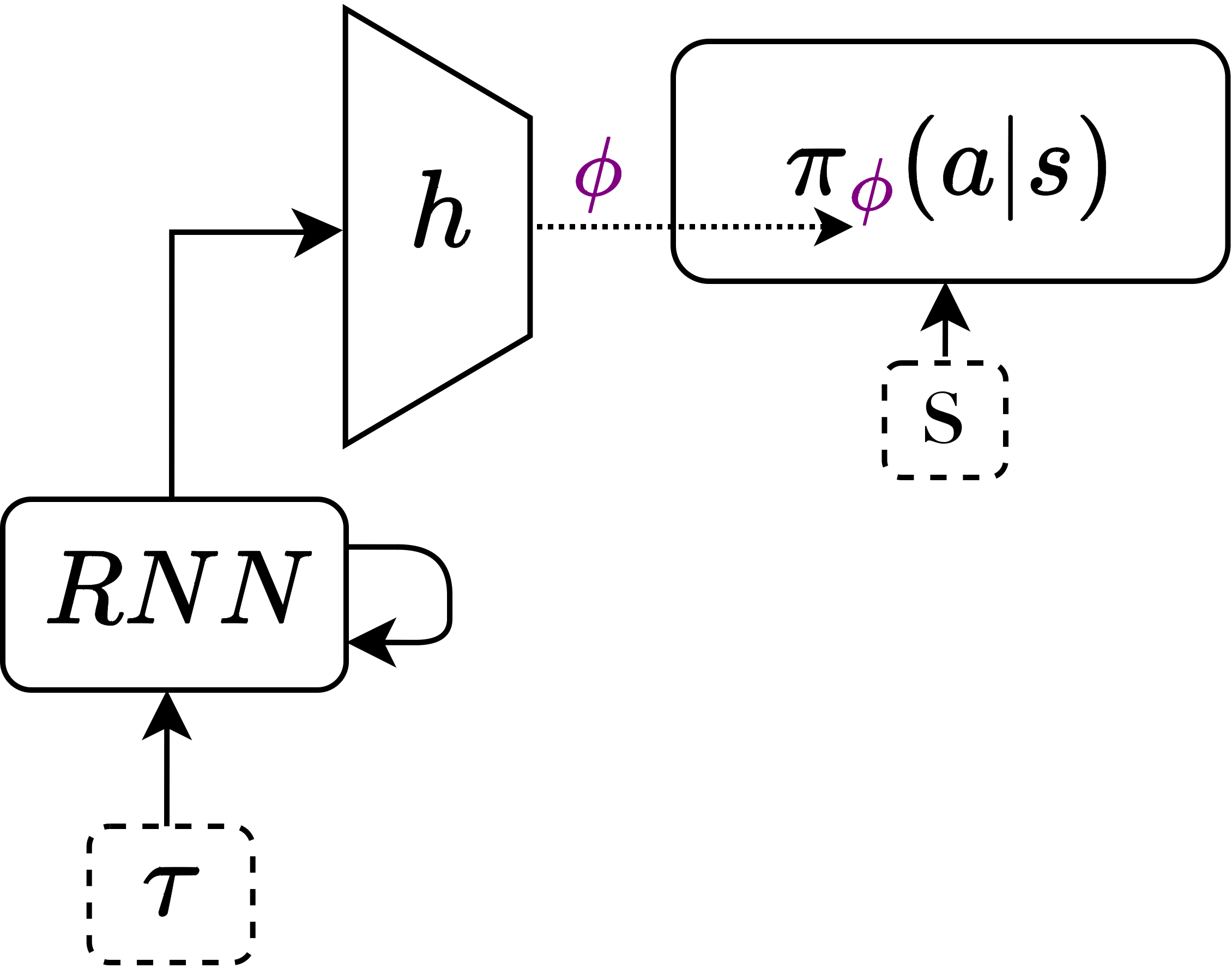}
        \caption{Hypernetwork}
        \label{fig:hyper}
    \end{subfigure}
    \hfill
    \begin{subfigure}[b]{0.32\textwidth}
        \centering
        \includegraphics[height=4.2cm]{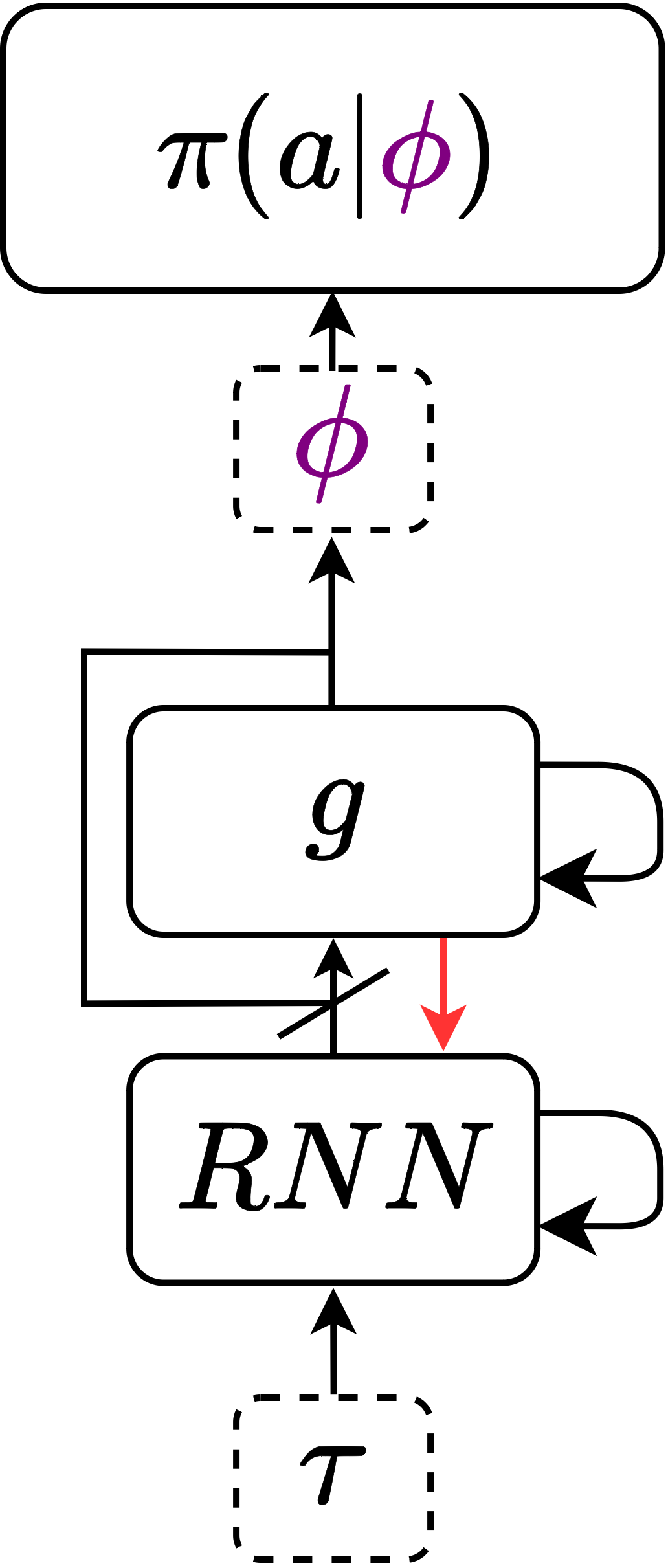}
        \caption{AMRL}
        \label{fig:amrl}
    \end{subfigure}
    \hfill
    \begin{subfigure}[b]{0.32\textwidth}
        \centering
        \includegraphics[height=4.2cm]{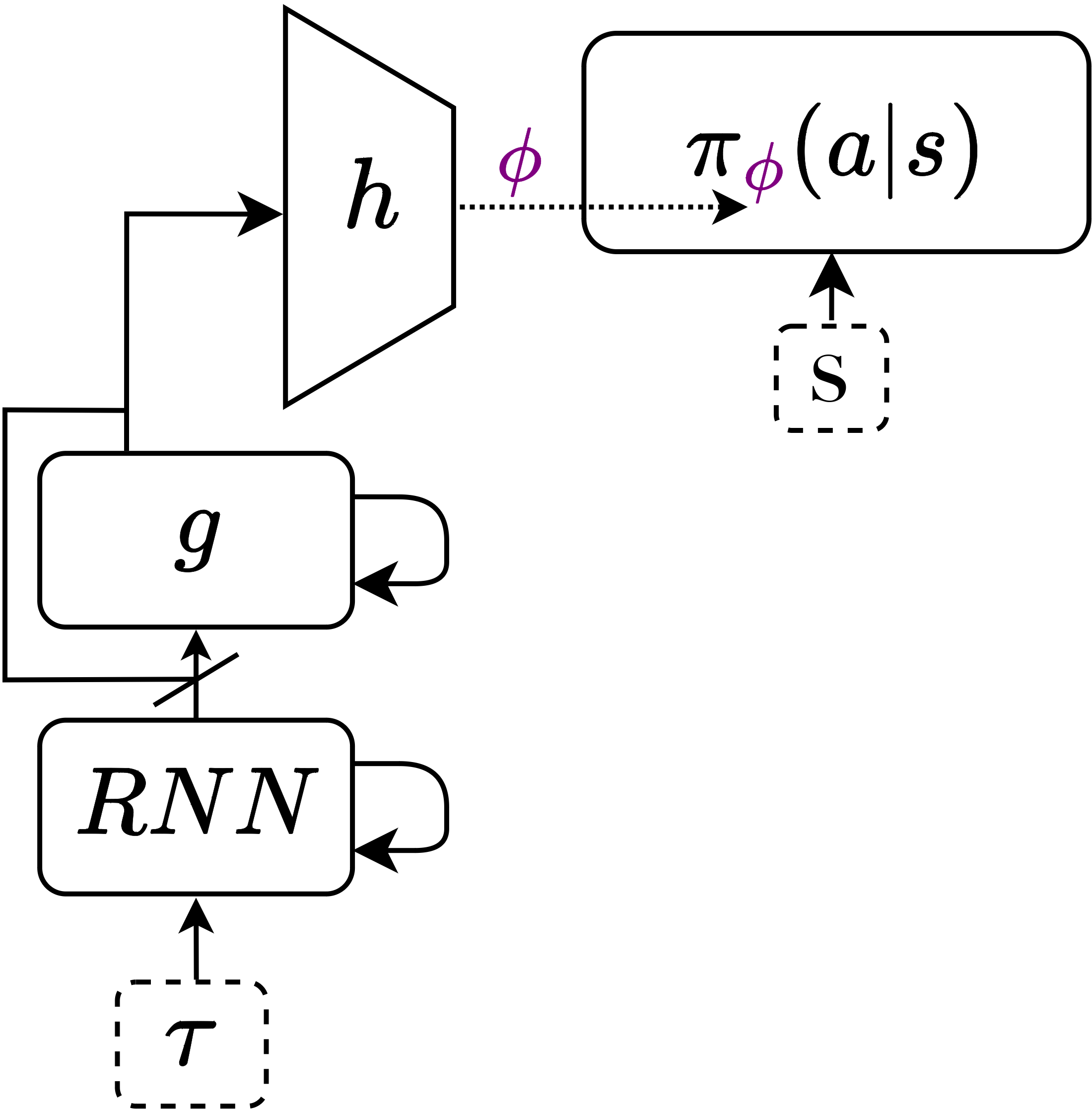}
        \caption{\method}
        \label{fig:splag}
    \end{subfigure}
    \caption{The hypernetwork from \citet{beck2023recurrent} is depicted in \ref{fig:hyper}, the AMRL model from \citet{beck2020AMRL} is depicted in \ref{fig:amrl}, and \method is depicted in \ref{fig:splag}. The angled line indicates a split connection that divides the neurons in half. The red arrow indicates a modified gradient computation in the backward pass. A hypernetwork is indicated by $h$. \method makes use of the hypernetwork architecture combined with the AMRL sequence model. The hypernetwork architecture is necessary for performant end-to-end training. Critically, \method also removes the gradient modification from AMRL which we show to be deleterious to performance.
    }
    \label{fig:architecture}
\end{figure}

\subsection{AMRL}
AMRL, depicted in Figure \ref{fig:architecture}, is a method for POMDPs that combines both permutation variant and permutation invariant components \citep{beck2020AMRL}.
While AMRL uses permutation invariant aggregators, the encoded inputs to the aggregators themselves are a function of history:
\begin{equation}
    e_t = RNN_\theta(\tau_1,...,\tau_t).
\end{equation}
Additionally, the neurons of each encoded input are split in half before aggregation, into $e_{t,1/2}$ and $e_{t,2/2}$.
The first half of the neurons are aggregated, while the second half skip the aggregation.
The complete sequence model is defined as
\begin{equation}
    f_\theta(\tau) = \textrm{concatenate}(e_{t,1/2};g_\theta(e_{1,2/2},...,e_{t,2/2})).
\end{equation}
Here, the RNN is able to handle short permutation variant sequence, which can then be integrated without respect to order by the permutation invariant aggregation over longer periods of time.
Since we split the neurons before aggregation, we call this process \textit{split aggregation}.

Additionally, AMRL modfies each Jacobian, $\frac{dg}{de_{i,2/2}} \forall_i$, when computing the chain rule in the backward pass.
Specifically, it overwrites the true Jacobian with the identity matrix, $I$.
This is called passing the gradient \textit{straight through}, or an \textit{ST gradient modification}.
Since the sum aggregator actually has $I$ as the true Jacobian, this can be seen as replacing the gradient with that of a sum.
For the average and max aggregators, the Jacobian is already similar to $I$.
Specifically, the average has a Jacobian that is $I/t$, and the max has an expected Jacobian that is $I/t$, under mild assumptions.
Thus, the modification can be seen as a rescaling of the gradient that does not diminish with time.
\citet{beck2020AMRL} hypothesize that the ST modification has no negative impacts while also preventing gradient decay.

\label{sec:problem}

\section{\method}
\label{sec:model}

\begin{figure}[t]
    \centering
    \begin{subfigure}[b]{0.42\textwidth}
        \centering
        \includegraphics[width=\textwidth]{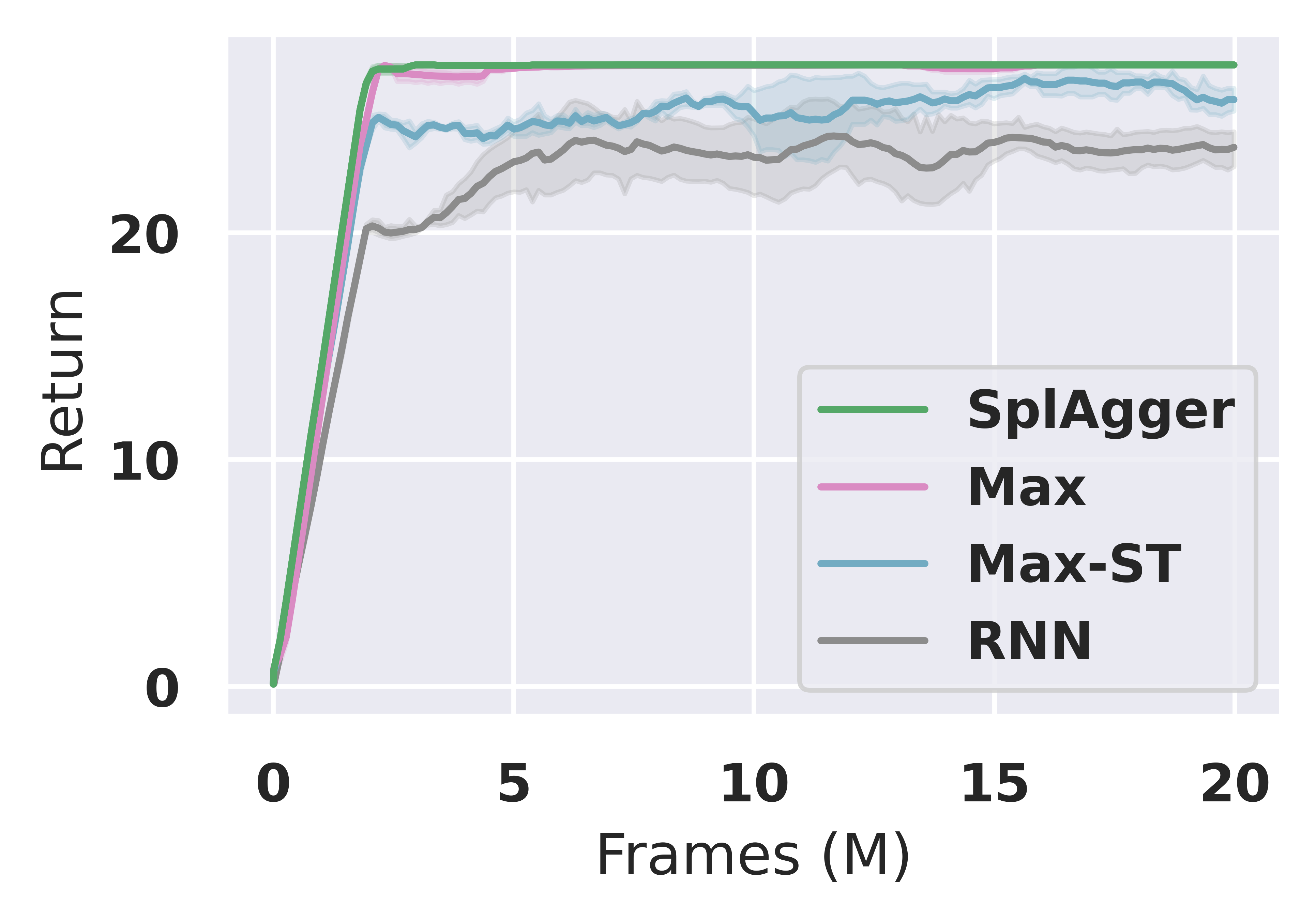}
        \caption{MC-LS}
        \label{fig:preview-mc}
    \end{subfigure}
    \hfill
    \begin{subfigure}[b]{0.42\textwidth}
        \centering
        \includegraphics[width=\textwidth]{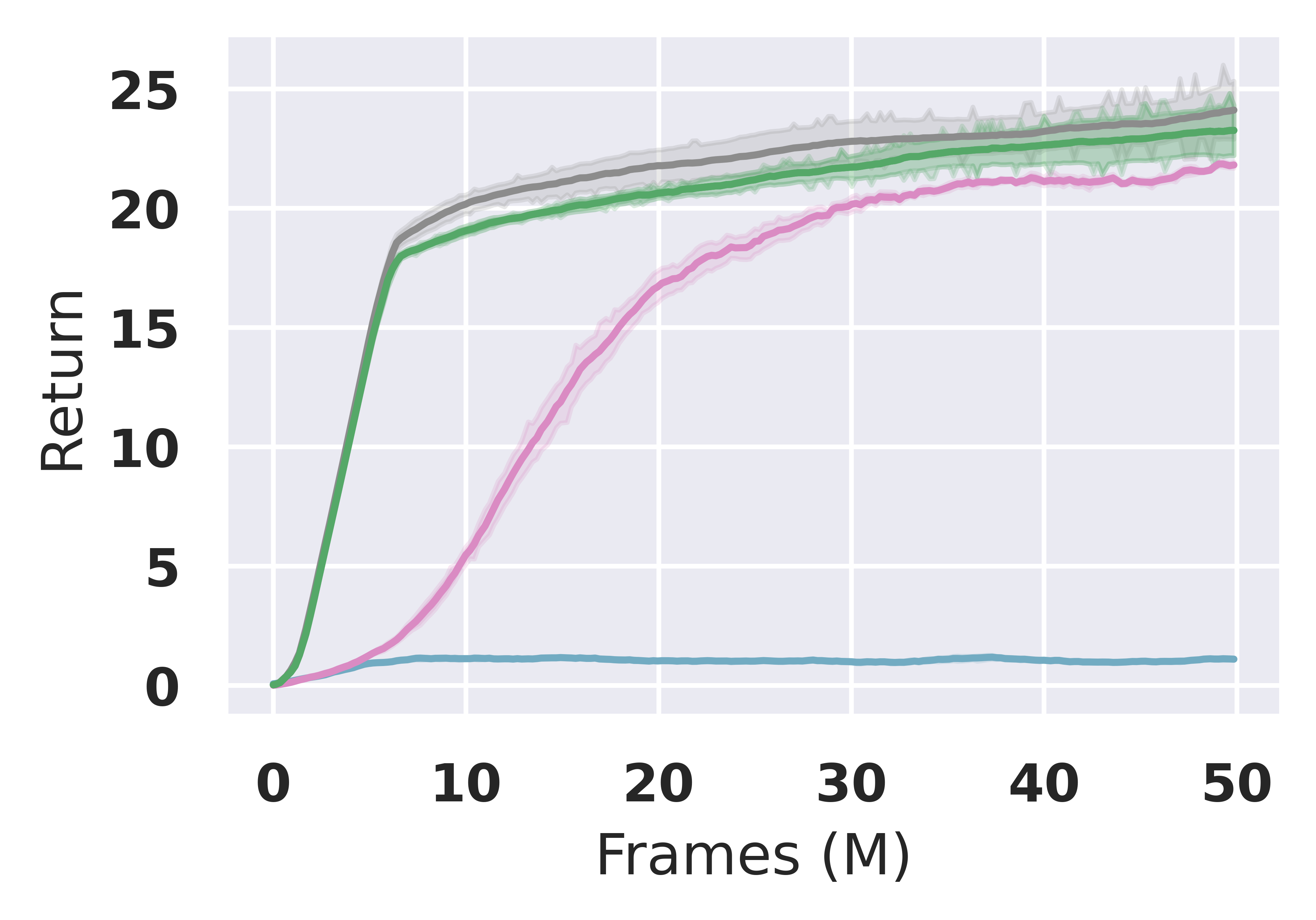}
        \caption{Planning Game}
        \label{fig:preview-pg}
    \end{subfigure}
    \caption{A preview of later results. The permutation invariance of the max aggregator improves returns relative the RNN on the MC-LS environment \citep{beck2020AMRL}, but decreases returns on the Planning Game \citep{ritter2021rapid}. Additionally, the ST gradient decreases the returns of the max aggregation. These results motivate \method, which achieves the highest returns. (Results are reported with a 68\% confidence interval, computed through bootstrapping with 1,000 iterations across three seeds, consistent with all plots presented.)
    }
    \label{fig:preview}
\end{figure}

Here we present our model, Split Aggregator, or \method.
As motivation, we first present a preview of our experimental results.
In Figure \ref{fig:preview}, we experiment with the permutation invariant model using a point-wise maximum aggregator, as suggested in AMRL, and a permutation variant model, the RNN.
Permutation variance improves performance on the some domains (Figure \ref{fig:preview-mc}), but decreases performance on others (Figure \ref{fig:preview-pg}).
Moreover, when we add the ST gradient modification, performance decreases severely.
These result motivate the need for our model, \method, which achieves the highest returns in both domains.

\method uses the same split aggregation as in AMRL, but without the ST gradient modification.
Simplifying AMRL by removing this gradient modification is key to its performance.
While the ST modification does prevent one type of gradient decay, permutation invariant aggregators already address the relevant type of gradient decay.
Moreover, the ST aggregator causes both the explosion of other gradients and a severe decrease in performance.
We show that this is the case in Sections \ref{sec:exp} and \ref{sec:analysis}, motivating the need for \method.
Additionally, \method uses a different architecture for the policy than in AMRL.
Recent results show that the hypernetwork \citep{ha2017hypernetworks} architecture is critical in unlocking the performance of end-to-end objectives and enabling black box methods to outperform task inference methods \citep{beck2023recurrent}.
Thus, the main idea behind \method is to add AMRL to hypernetworks trained end-to-end but remove the ST gradient modification.
\method is the combination of these components and is depicted in Figure \ref{fig:splag}.

\section{Experiments}
\label{sec:exp}

In this section we evaluate \method on several domains.
First, we evaluate on two standard meta-RL benchmarks in Section \ref{sec:mujoco}, to make sure that the aggregation method does not harm performance on environments without large demands on the sequence models.
Second, we evaluate on two prior meta-RL benchmarks designed to test sequence models in mazes in Section \ref{sec:memexp}.
We additionally evaluate on three environments design to systematically test different components of \method in Section \ref{sec:abl}.

On the four primary benchmark environments, we compare to four baselines.
Hyperparamter tuning is detailed in Appendix \ref{sec:tuning}.
Since prior results demonstrate the need for hypernetworks when training end-to-end \citep{beck2023recurrent}, all baselines have been evaluated using hypernetworks, with design choices detailed in Appendix \ref{sec:tuning}.
We additionally present negative results on a novel initialization method in Appendix \ref{sec:invinit}.
The baselines evaluated primarily differ in their choice of aggregation function, $g$, and encoding of inputs, $e_t$. 
The baselines are described below.

\paragraph{RNN.} The RNN baseline can be written $f(\tau)=e_t=RNN(\tau_1,...,\tau_t)$. Here there is no aggregation function, $g$, and the baseline uses a standard gated recurrent unit \citep{cho2014learning}, as in \citet{zintgraf2021varibad} and \citet{beck2023recurrent}. 

\paragraph{CNP.} The conditional neural process (CNP) consists of permutation invariant aggregation without any additional components \citep{garnelo2018conditional}.
Specifically, $f(\tau)=g(e_1,...,e_t)$.
Here, $e_t$ is a linear encoding of $\tau_t$.
We use the mean operator for $g$, as suggested by \citet{garnelo2018conditional}.

\paragraph{AMRL.} AMRL \citep{beck2020AMRL} uses an RNN to encode $e_t$ in addition to permutation invariant aggregation, and is described in Section \ref{sec:background}. For AMRL, we use the pointwise maximum aggregator in our experiments, both to match the aggregator used in \method, and because that aggregator was found to be strongest by \citet{beck2020AMRL}.

\paragraph{PEARL.} The PEARL baseline uses the aggregation method from the PEARL algorithm \citep{rakelly2019efficient}, which incorporates permutation invariance into the probability density function of a stochastic latent variable summarizing history. 
The density function is modelled as a product over individual transitions in the data.
We can write this as $f(\tau)= g(e_1,...,e_t) = z \sim \alpha \prod_{t = 1}^{t=T} \mathcal{N}(z;\mu_t=e_{t,1/2},\sigma^2_t=diag(e_{t,2/2}))$, where $e_t$ is a linear encoding of $\tau_t$ and $\alpha$ is a normalizing constant.
To compare the effects of aggregation in isolation, our PEARL baseline only implements the aggregation method used in PEARL, and leaves the rest of the algorithmic choices the same as in \method.
Additional design choices and hyperparameters for PEARL are presented in Appendices \ref{sec:tuning} and \ref{sec:pearl_additional}.

\subsection{MuJoCo Benchmarks}
\label{sec:mujoco}

The first two environments for benchmarking are variants of MuJoCo proposed by \citet{zintgraf2021varibad}, and both involve legged locomotion.
While these environments have no great demands on memory, they are common meta-RL benchmarks \citep{humplik2019meta,rakelly2019efficient,zintgraf2021varibad,beck2022hyper,beck2023recurrent} that enable us to evaluate what effect \method has on standard RL tasks.
See Appendix \ref{sec:env_details} for details on the environments.

\begin{figure}[t]
    \centering
    \begin{subfigure}[b]{0.45\textwidth}
        \centering
        \includegraphics[width=\textwidth]{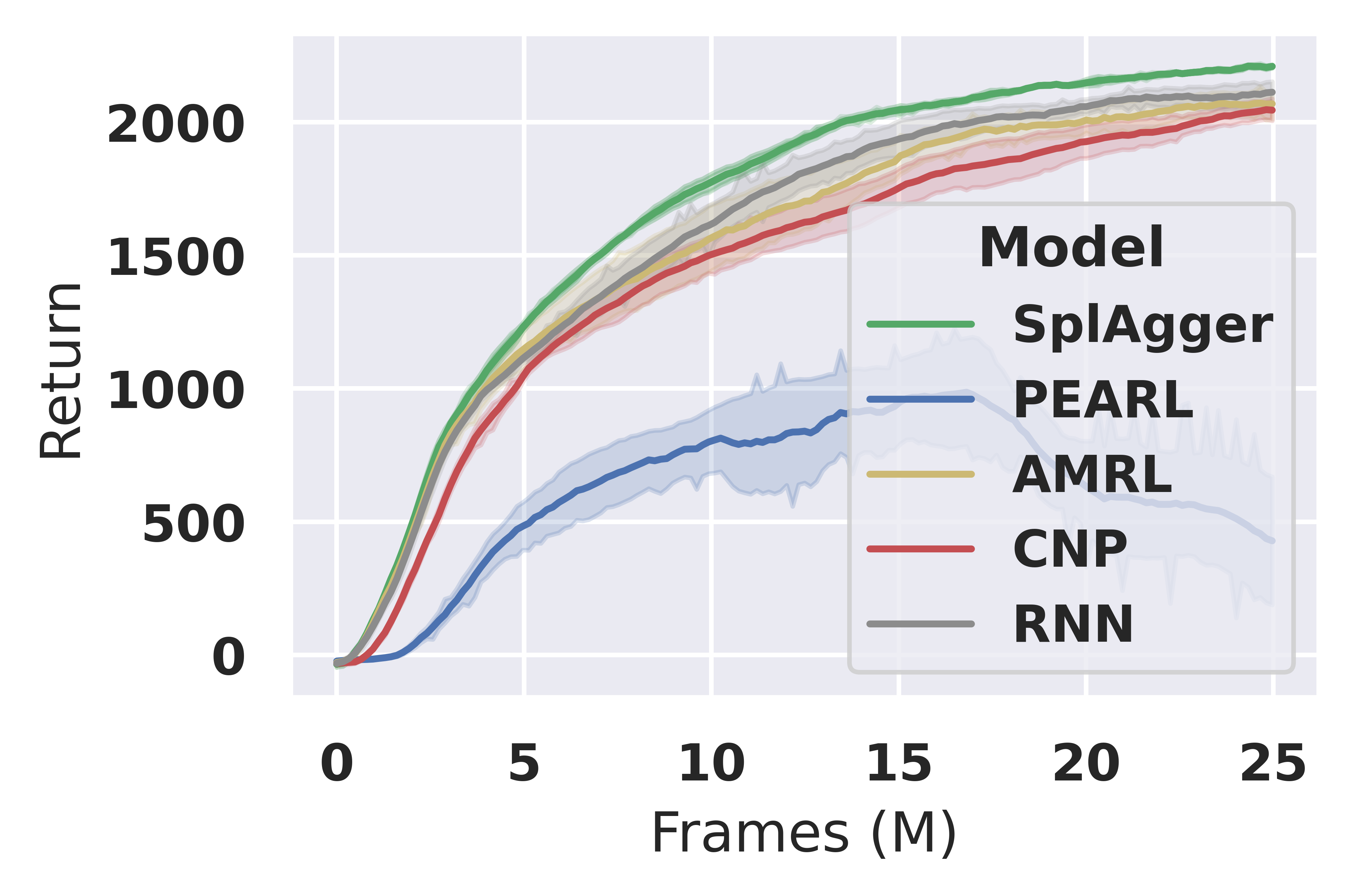}
        \caption{Cheetah-Dir}
        \label{fig:sub1}
    \end{subfigure}
    \hfill
    \begin{subfigure}[b]{0.45\textwidth}
        \centering
        \includegraphics[width=\textwidth]{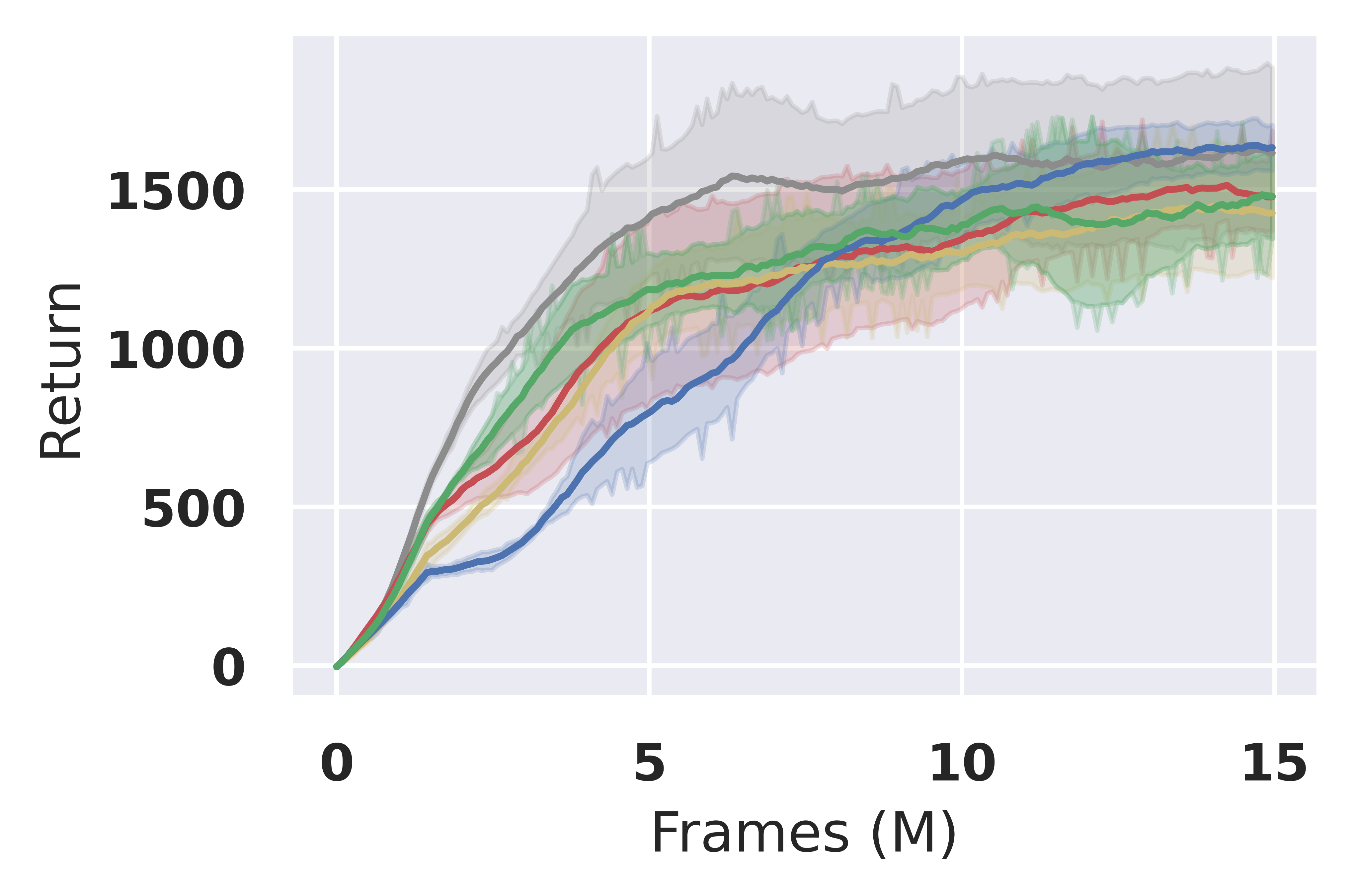}
        \caption{Walker}
        \label{fig:sub2}
    \end{subfigure}
    \caption{Results on MuJoCo benchmarks. \method achieves the same or better results on both domains. PEARL achieves significantly lower return on Cheetah-Dir.
    }
    \label{fig:mujoco}
\end{figure}

Results are shown in Figure \ref{fig:mujoco}.
\method achieves the greatest return, though the improvement is modest.
Training the PEARL baseline on this domain is unstable and PEARL receives significantly lower returns.
On Walker, we see similar performance across all methods.
Overall, \method achieves similar or greater returns compared to other baselines.
This demonstrates that our method, designed to improve environments with difficult demands on memory, also does not decrease performance on domains with limited memory requirements.

\begin{figure}[t]
    \captionsetup[subfigure]{justification=centering}
    \centering
    \begin{subfigure}[b]{0.45\textwidth}
        \centering
        \includegraphics[trim=110 110 250 70, clip, width=\textwidth]{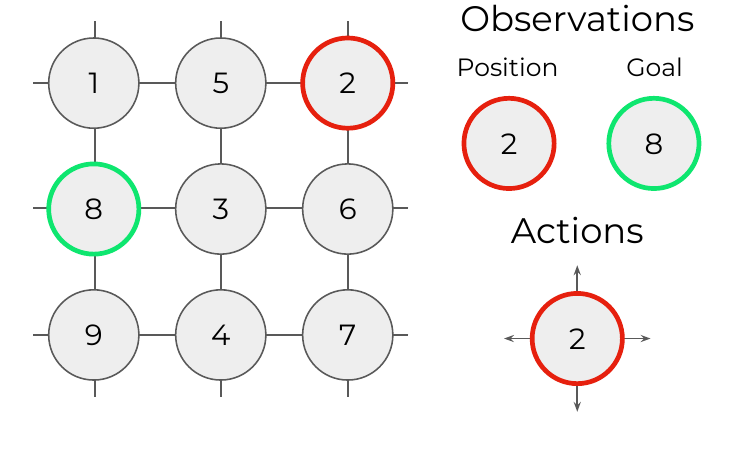}
        \caption{Planning Game}
        \label{fig:planning_env}
    \end{subfigure}
    \hfill
    \begin{subfigure}[b]{0.155\textwidth}
        \centering
        \includegraphics[width=\textwidth]{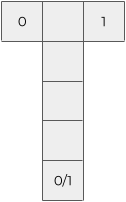}
        \caption{T-LS \newline (T-Maze)}
        \label{fig:tmaze_env}
    \end{subfigure}
    \hfill
    \begin{subfigure}[b]{0.155\textwidth}
        \centering
        \includegraphics[width=\textwidth]{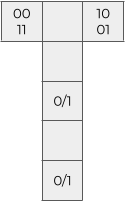}
        \caption{T-Maze \newline Agreement}
        \label{fig:tmaze_agree_env}
    \end{subfigure}
    \hfill
    \begin{subfigure}[b]{0.155\textwidth}
        \centering
        \includegraphics[width=\textwidth]{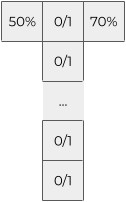}
        \caption{T-Maze \newline Latent}
        \label{fig:fig:tmaze_latent_env}
    \end{subfigure}
    \caption{Depictions of the Planning Game, T-LS, T-Maze Agreement, and T-Maze Latent environments used in Sections \ref{sec:memexp} and 
    \ref{sec:abl}.
    }
    \label{fig:envs}
\end{figure}

\subsection{Memory Benchmarks}
\label{sec:memexp}

We additionally conduct tests on two environments, T-LS and MC-LS, proposed by \citet{beck2020AMRL}.
Both of these environments were designed to test long-term memory, and the latter has been used previously in meta-RL \citep{beck2023recurrent}. 
The T-LS environment is depicted in Figure \ref{fig:tmaze_env}.
The MC-LS environment is designed to challenge an agent's long-term memory based on visual cues from Minecraft. 
Environment details can be found in Appendix \ref{sec:env_details}.

Results in Figure \ref{fig:mem_results} show that \method achieves the highest sample efficiency on both environments.
The RNN and PEARL are not able to learn the optimal policy within the allotted number of frames.
The reasons for the failure of the RNN are discussed in Section \ref{sec:analysis}, while potential reasons for the failure of PEARL are analyzed in Appendix \ref{sec:pearl_analysis}.
While CNP is able to learn optimally, it requires significantly more frames on T-LS.
AMRL achieves similar performance.
Both AMRL and CNP are similar to our method, \method.
However, AMRL differs in its gradient estimation and CNP differs in its use of mean aggregator, instead of max, and its lack of an RNN.
To fully understand the contribution of each, we systematically modify these components in isolation, in the next section.

\subsection{Alternative Aggregation}
\label{sec:abl}

In this section we introduce modifications of \method, and three new environments designed to test these modifications.
Since the existing baselines differ in their choice of aggregation function, $g$, encoding of inputs, $e_t$, and use of the ST gradient modification, we systematically test these differences here. 
The most relevant additional baselines are described below, with the rest detailed in Appendix \ref{sec:alt_agg}.

\textbf{\method-noSplit} removes the split aggregation from \method, but still uses max aggregation and an RNN to encode $e_t$.
Comparing to this method allows us to validate the use of the split aggregation in \method. 
\textbf{\method-noRNN} removes the RNN from \method in order to test the effects of removing permutation variant components.
Since this obviates the need for the split connection, that component is removed as well.
Without these components, this method is equivalent to just computing a maximum over linear encodings of each transition, $\tau_t$.
\textbf{AMRL-noRNN} removes the RNN and split connection from AMRL.
Without these components, this method is equivalent computing a maximum over linear encodings of each transition, $\tau_t$, along with the ST gradient modification.
\textbf{\method-avg} replaces the max operator in \method with an average, in order to test the effects of alternative permutation invariant operators. \method with other operators (avgmax, softmax, wsoftmax) that interpolate between the average and max are evaluated as well, with details in Appendix \ref{sec:alt_agg}.

\textbf{Planning Game.} First, we evaluate on the Planning Game \citet{ritter2021rapid}, in order to evaluate the need for permutation variant components and how to combine them with permutation invariant components. 
This environment tests an agents ability to discover and remember multiple pieces of information required for subsequent navigation.
The Planning Game is useful for evaluation here due to the existence of both a permutation invariant optimal policy and permutation invariant suboptimal policy.
The environment is depicted in Figure \ref{fig:planning_env} and detailed in Appendix \ref{sec:env_details}.

\begin{figure}[t]
    \begin{subfigure}[b]{0.32\textwidth}
        \includegraphics[width=.4\textwidth,right]{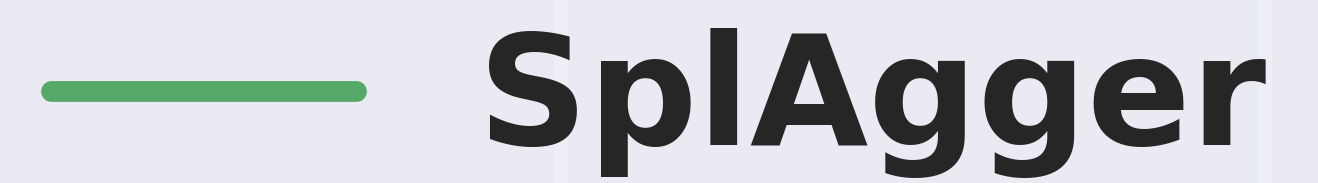}
    \end{subfigure}
    \begin{subfigure}[b]{0.32\textwidth}
        \centering
        \includegraphics[width=.4\textwidth]{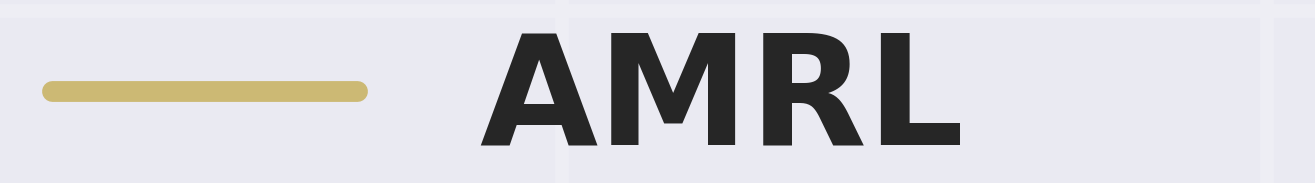}
    \end{subfigure}
    \begin{subfigure}[b]{0.32\textwidth}
        \includegraphics[width=.4\textwidth,left]{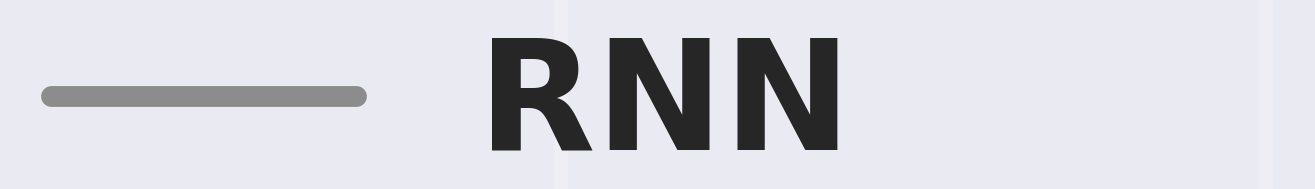}
    \end{subfigure}
    \centering
    \begin{subfigure}[b]{0.32\textwidth}
        \centering
        \includegraphics[width=\textwidth]{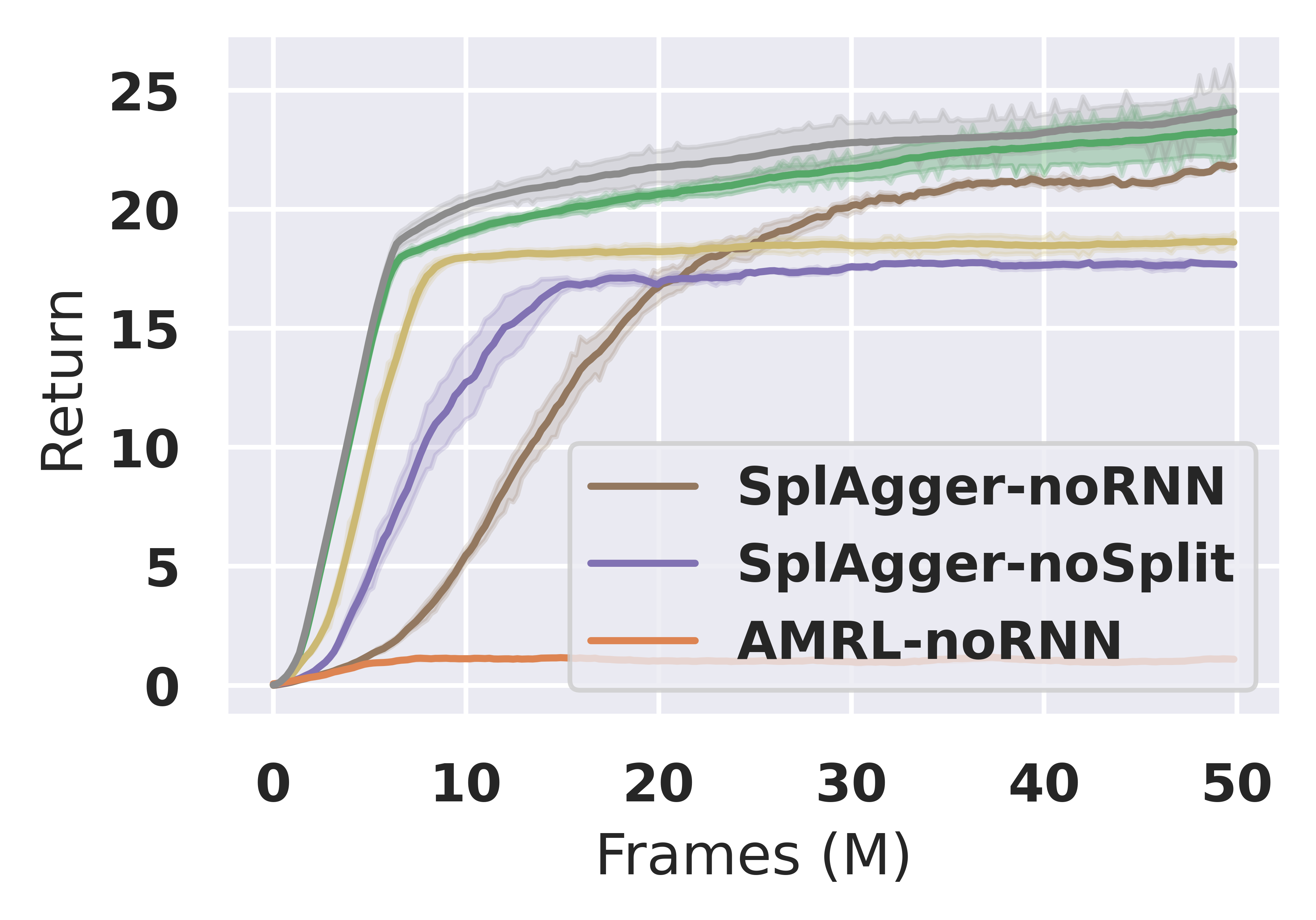}
        \caption{Planning Game}
        \label{fig:plan}
    \end{subfigure}
    \hfill
    \begin{subfigure}[b]{0.32\textwidth}
        \centering
        \includegraphics[width=\textwidth]{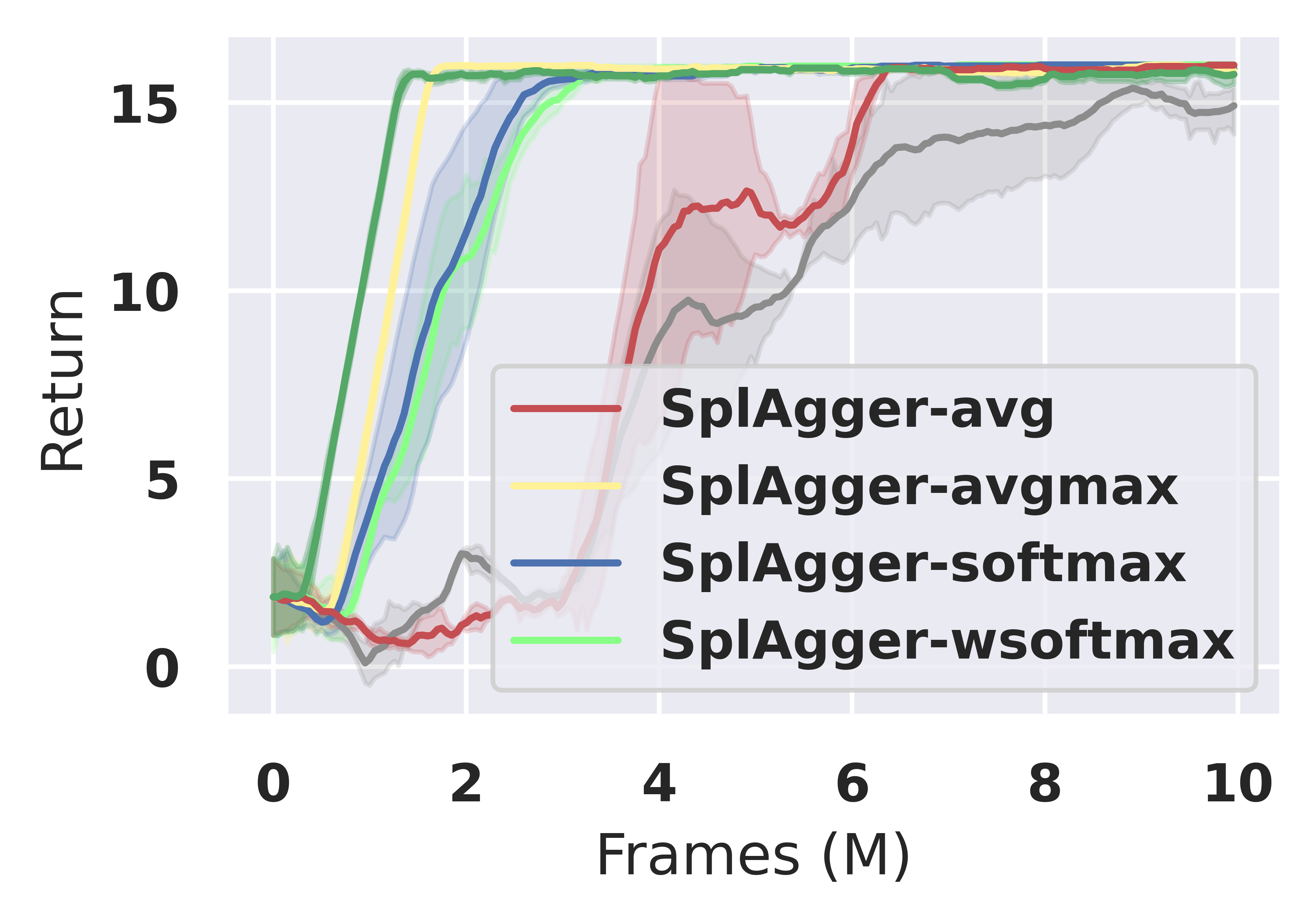}
        \caption{T-Maze Agreement}
        \label{fig:ta}
    \end{subfigure}
    \hfill
    \begin{subfigure}[b]{0.32\textwidth}
        \centering
        \includegraphics[width=\textwidth]{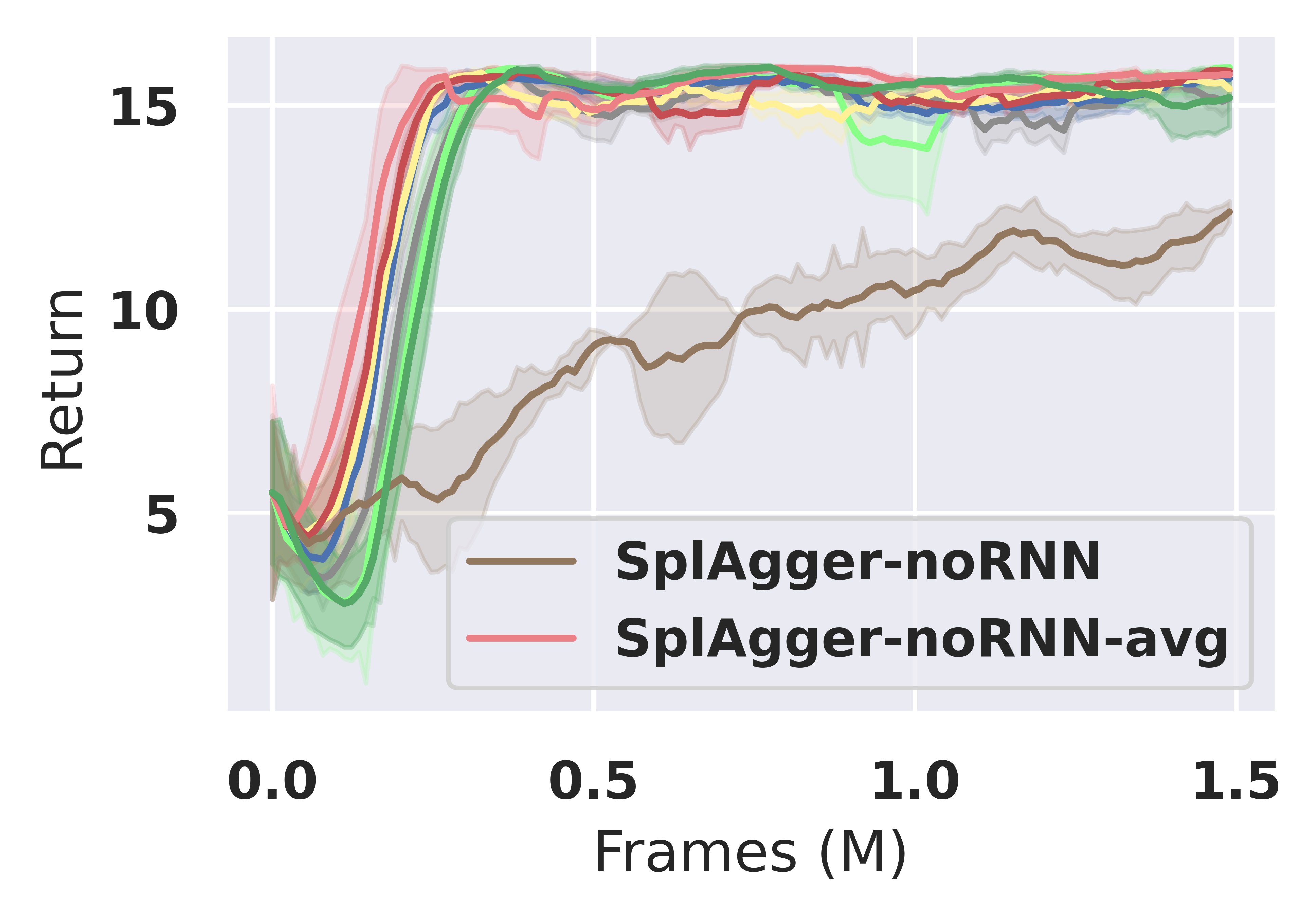}
        \caption{T-Maze Latent}
        \label{fig:tld}
    \end{subfigure}
    \caption{\method achieves returns that are equal to or higher than other methods.
    The Planning Game shows the importance of incorporating RNNs, either in isolation or with split aggregation, and the failure of gradient modification as in AMRL.
    Combining the RNN and permutation invariant aggregation, without the split connection (\method-noSplit), decreases performance of each.
    The T-Maze Agreement domain shows the max operator to be beneficial, enabling performance even greater than the RNN when used with \method.
    The T-Maze Latent environment shows that \method is able to make the max aggregator performant, even in environments where the computing the average alone is superior.
    Note that each legend shows the additional methods introduced for that experiment, while the legend at the top shows methods from prior experiments.
    }
    \label{fig:subfigureExample3}
\end{figure}

On this domain, we evaluate methods that modify how the RNN is combined with the permutation invariant aggregation.
Results in Figure \ref{fig:plan} show that \method and RNN learn the fastest in this domain.
Both AMRL and \method without the split connection learn a suboptimal policy.
This demonstrates the detrimental effects of the AMRL gradient modification and the benefit of the split aggregation, which motivates \method.
We also see that AMRL without an RNN fails to learn any reasonable policy, achieving near zero reward.
Since the only difference between this and \method without an RNN is the use of the ST gradient modification, this shows strong evidence of the detrimental effects of the gradient modification in AMRL.
We analyze the causes in Section \ref{sec:analysis}.
While both \method with an RNN and \method without an RNN outperform the sub-optimal exploration policy eventually, and achieve similar returns ultimately, \method with an RNN learns faster initially.
We discuss this further in Section \ref{sec:analysis}.

\begin{figure}[t]
    \centering
    \begin{subfigure}[b]{0.45\textwidth}
        \centering
        \includegraphics[width=\textwidth]{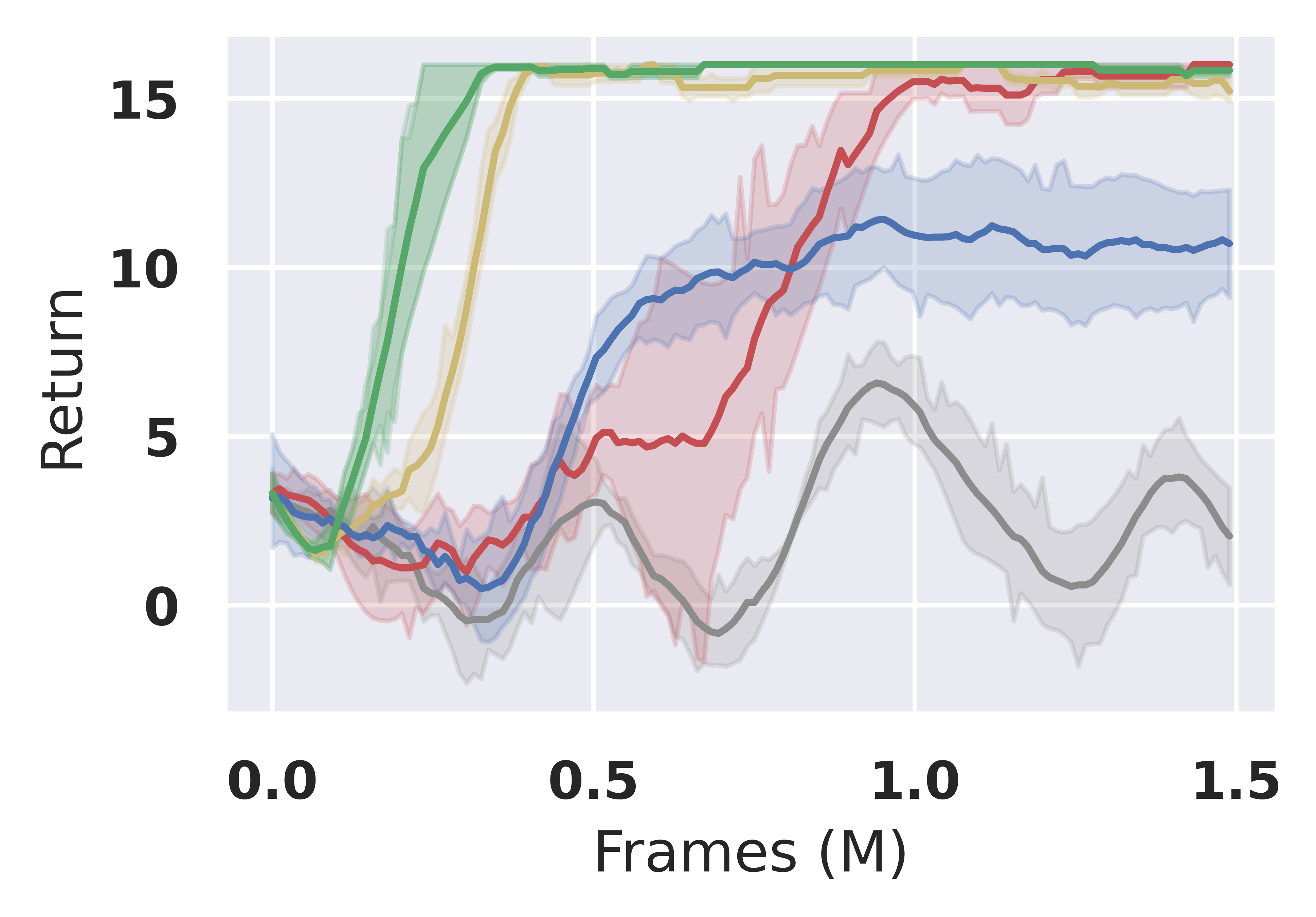}
        \caption{T-LS}
    \end{subfigure}
    \hfill
    \begin{subfigure}[b]{0.45\textwidth}
        \centering
        \includegraphics[width=\textwidth]{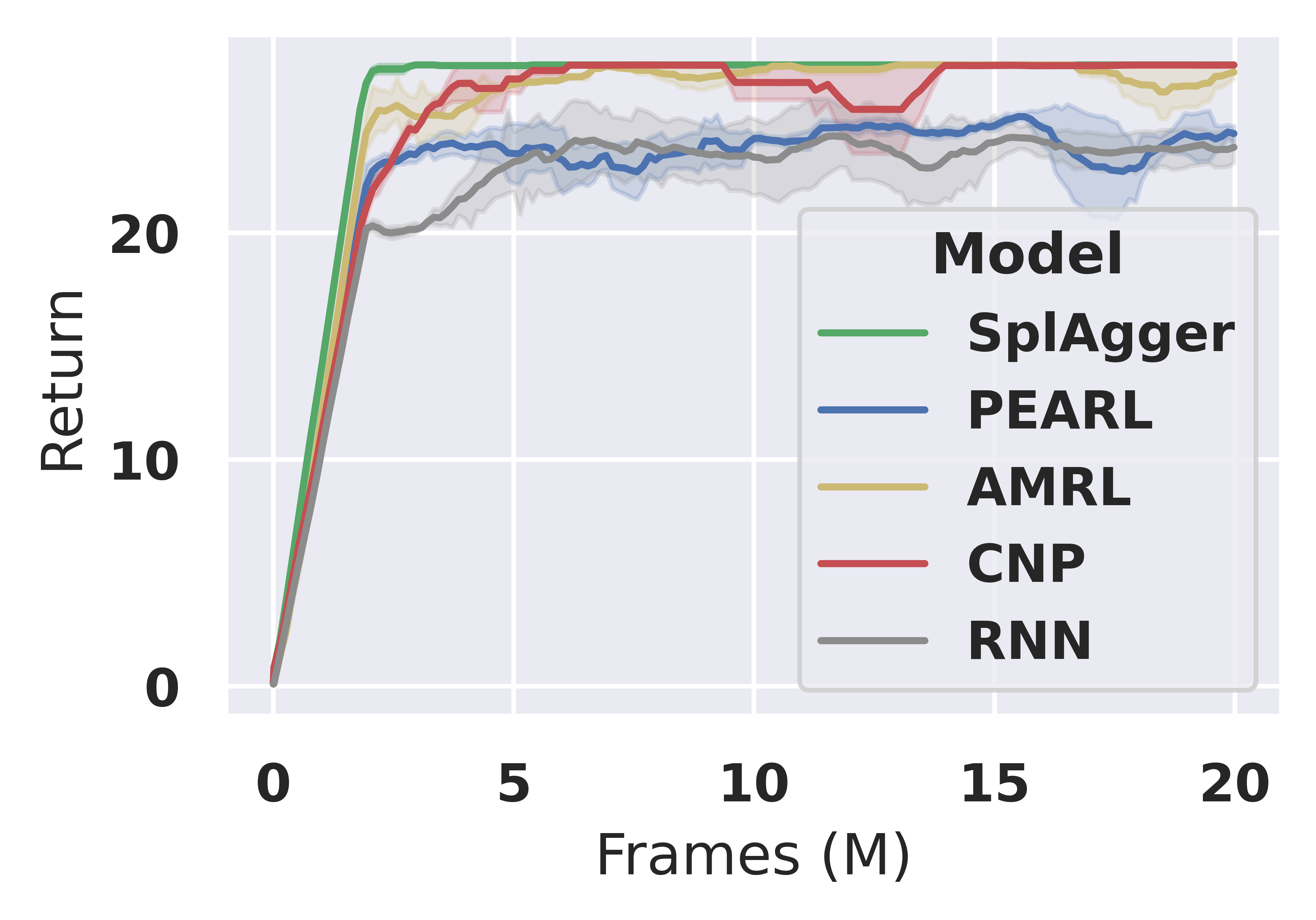}
        \caption{MC-LS}
    \end{subfigure}
    \caption{Results on memory benchmarks. \method achieves the highest returns on both domains, indicating the fastest learning. The standard RNN is not able to learn on either domain within the allotted number of frames.}
    \label{fig:mem_results}
\end{figure}

\paragraph{T-Maze Agreement.} Second, we evaluate on T-Maze Agreement in order to investigate the specific permutation invariant operator, $\bigoplus$.
In this environment, the agent receives two binary signals: one at the beginning of the maze and one in the middle.
The agent must open a door depending on whether the signals agree or disagree.
This environment is depicted in Figure \ref{fig:tmaze_agree_env}.
While the max aggregation can easily identify the support of the state distribution in the data \citep{beck2020AMRL}, and thus easily identify which signals have been seen in each state, the average aggregation must learn to adjust the representation of all states to interpret the average.
Thus, we hypothesize that this environment is easier for max aggregation and harder for mean aggregation.

On this domain, we evaluate methods that modify the specific permutation invariant operators.
Results in Figure \ref{fig:ta} show that \method achieves the highest returns, demonstrating the superiority of max aggregation in this environment.
While the RNN and average variant of \method are able to learn, they require more frames, as expected.
The avgmax, softmax, and wsoftmax all learn almost as quickly as \method.
Thus, we see that \method is fairly robust to the choice of operator in this domain, as long as it computes some information about the maximum.

\paragraph{T-Maze Latent.} Finally, we evaluate on T-Maze Latent, which also modifies the T-Maze environment in order to investigate the specific permutation invariant operator.
In this environment, the agent receives an indicator at every timestep.
This indicator is either drawn from $\{0,1\}$ with a 50\% chance of 1 or a 70\% chance of 1, depending on the task.
This environment is depicted in Figure \ref{fig:fig:tmaze_latent_env}.
The average aggregator should quickly reveal the latent variable, as the variance of the mean decreases, whereas the max aggregator should have a more difficult time counting the occurrence of indicators in different states.
Thus, we hypothesize that this environment should be harder for max aggregation methods and easier for average aggregation.

On this domain, as on T-Maze Agreement, we evaluate methods that modify the specific permutation invariant operators.
Results in Figure \ref{fig:tld} show that, surprisingly, all operators used with \method learn at approximately the same rate.
We hypothesize that here, \method is able to fall back upon leveraging the RNN.
Since the environment was designed to be more difficult for the max operator, we predict that there may be a difference between the operators when the RNN and skip connection of \method are not used.
To test this hypothesis, we conduct two additional experiments.
We test both \method without an RNN, but still with the default max operator (\method-noRNN), and \method without an RNN but with the average operator (\method-noRNN-avg).
Since the removal of the RNN obviates the need for the split aggregation, the split aggregation is removed as well.
We see that, as predicted, the method with the average operator performs better than the method with the max operator, when the RNN and split connection are removed.
This demonstrates that the combination of the RNN and split aggregation make \method remarkably robust to different environments, and justifies both our aggregation method and the max operator.

\section{Analysis}
\label{sec:analysis}
In this section we analyze different sequential models to gain insights into their performance.
We investigate why RNNs remain useful in some cases, even when permutation invariance should be sufficient, and why AMRL performs poorly in our experiments.
Specifically, we find permutation variance to be useful when there exist permutation variant suboptimal policies that form a useful stepping stone for learning optimal policies later.
Additionally, we find that both AMRL and \method prevent certain types of gradient decay, but AMRL also causes other gradients to explode.

\subsection{Learning Suboptimal Policies}

While sensitivity to permutation is not required to learn optimal policies in meta-RL, we find that the RNN surprisingly improves sample efficiency on the Planning Game.
As discussed in Section \ref{sec:exp}, 
the Planning Game has a permutation variant suboptimal policy.
This policy re-explores all states after every goal is reached.
To do this, the agent must remember where it is in a sequence of exploratory actions, and then restart the sequence when a new goal is found.
We examine rollouts to confirm that \method first learns this suboptimal policy.
While \method without an RNN can surpass the suboptimal policy eventually, it also achieves achieves lower returns throughout training compared to an RNN, since the RNN learns the suboptimal policy faster.
Thus, RNNs can achieve higher returns sooner, when there is a permutation variant suboptimal policy that can act as a stepping stone for learning the optimal policy.
We hypothesize that the reason \method without an RNN cannot learn the suboptimal policy is that the max aggregation can record that a goal has been reached, but cannot identify when that goal was reached, or that it was \textit{just} reached.

\begin{figure}
    \centering
    \includegraphics[width=.5\textwidth]{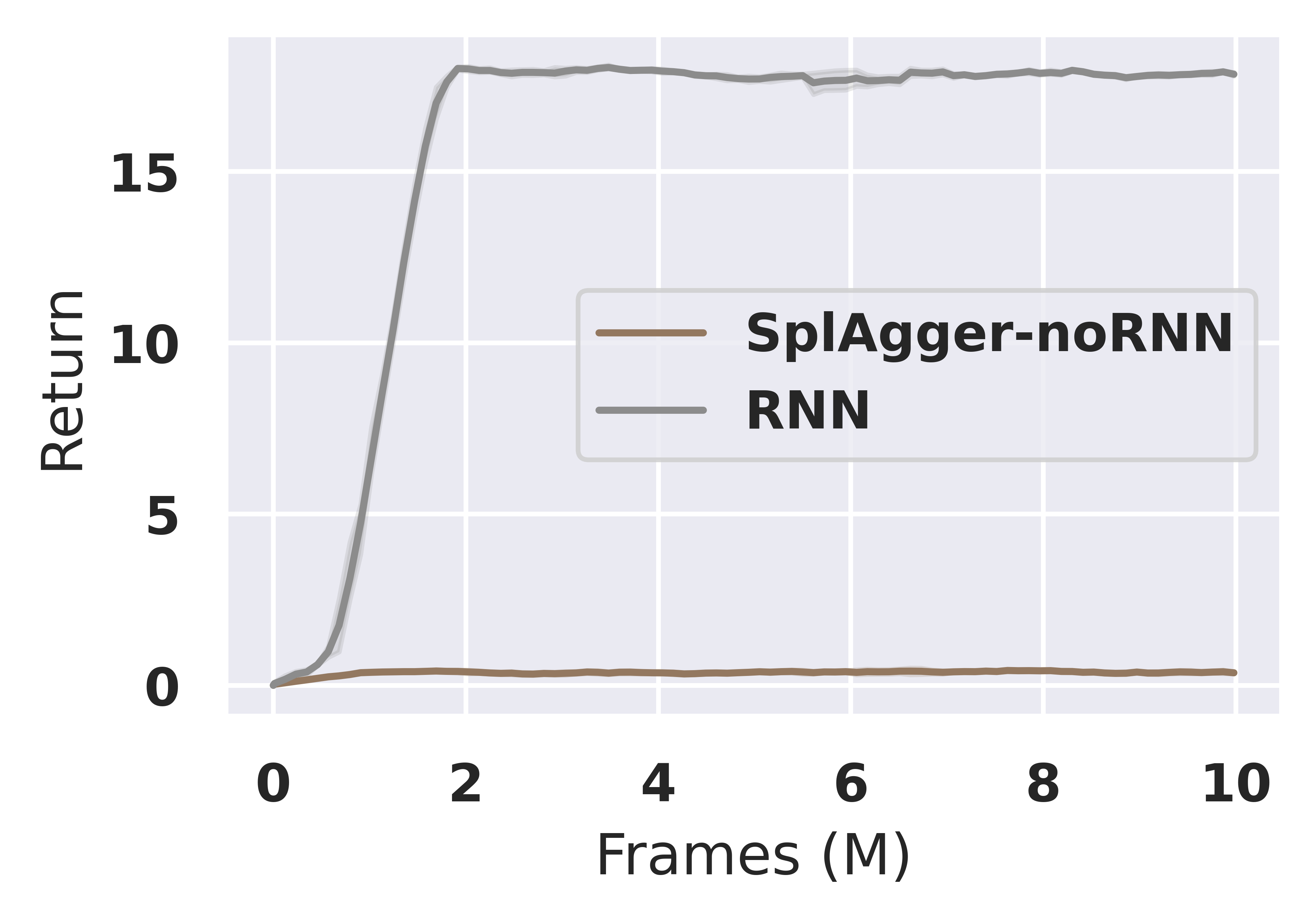}
    \caption{Here we show results on a modified Planning Game. If no observation is given to the agent, an RNN is required to learn the proper exploration strategy. This is the same exploration as required by the easier sub-optimal policy in the original Planning Game.}
    \label{fig:pj}
\end{figure}

To confirm this, we perform an additional experiment, in which the agent receives no observation of the state, creating partial observability.
All it observes is whether it is currently at the goal state.
Hence, there is no way for it to distinguish which MDP it is in, and the problem can no longer be modelled as a distribution of MDPs.
In this case, the observations are not Markov and permutation invariant aggregation no longer suffices for decision making.
Still, the optimal policy in this environment requires the same strategy as the suboptimal policy in the Planning Game: explore every state until the goal is reached.
When the agent reaches a goal, the goal location is reset, and the agent must restart exploration.
Figure \ref{fig:pj} shows that the RNN is able to learn this policy, while \method without an RNN is not.
This shows how, even in a distribution of MDPs, where the Markov property holds in each MDP, permutation variance can improve sample efficiency, due to the presence of non-Markov policies that are suboptimal but faster to learn.

\subsection{Gradient Decay}

\begin{figure}[t]
    \centering
    \begin{subfigure}[b]{0.45\textwidth}
        \centering
        \includegraphics[width=\textwidth]{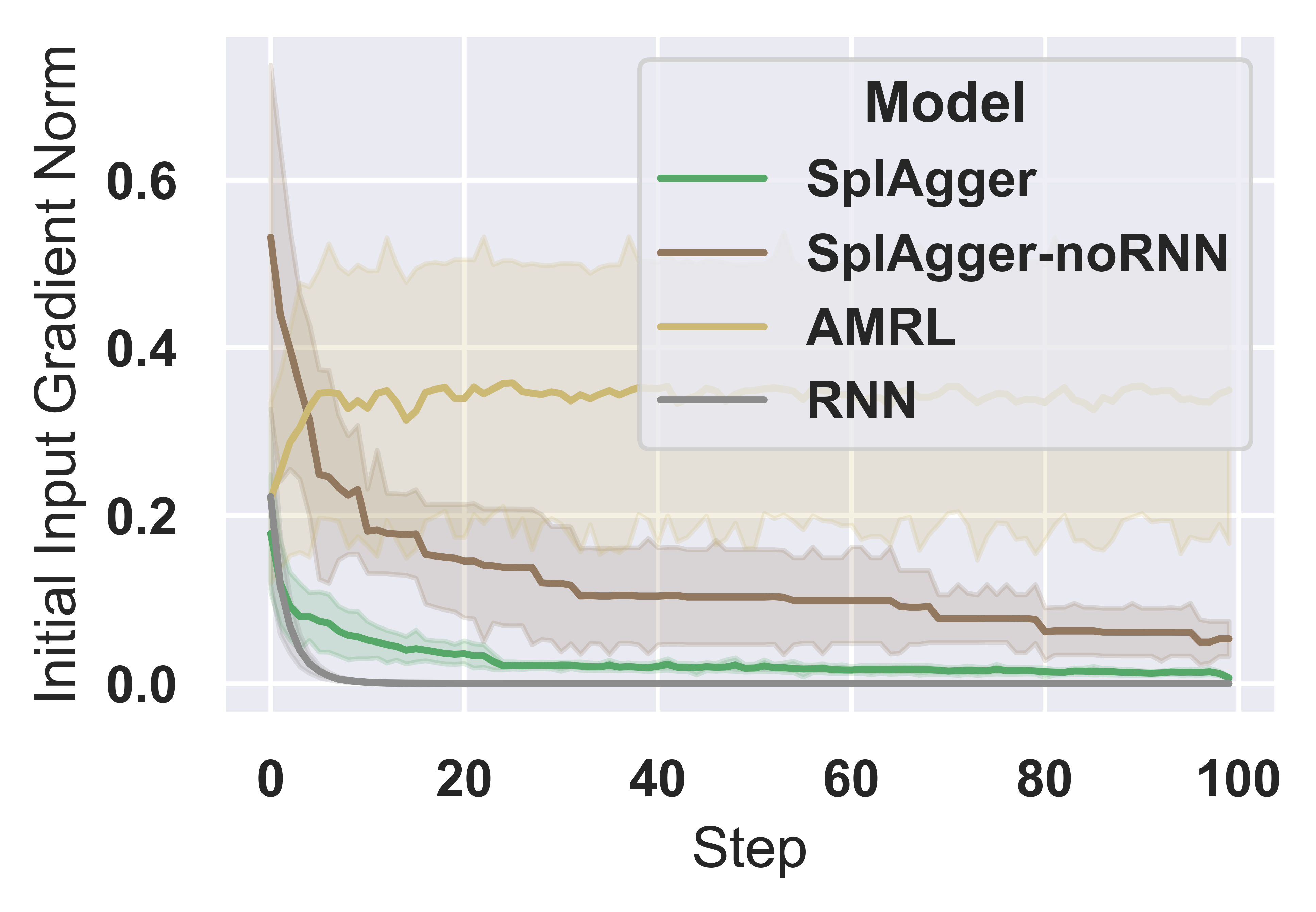}
        \caption{Gradient w.r.t. Initial Transition Over Time}
        \label{fig:amrl_grad}
    \end{subfigure}
    \hfill
    \begin{subfigure}[b]{0.45\textwidth}
        \centering
        \includegraphics[width=\textwidth]{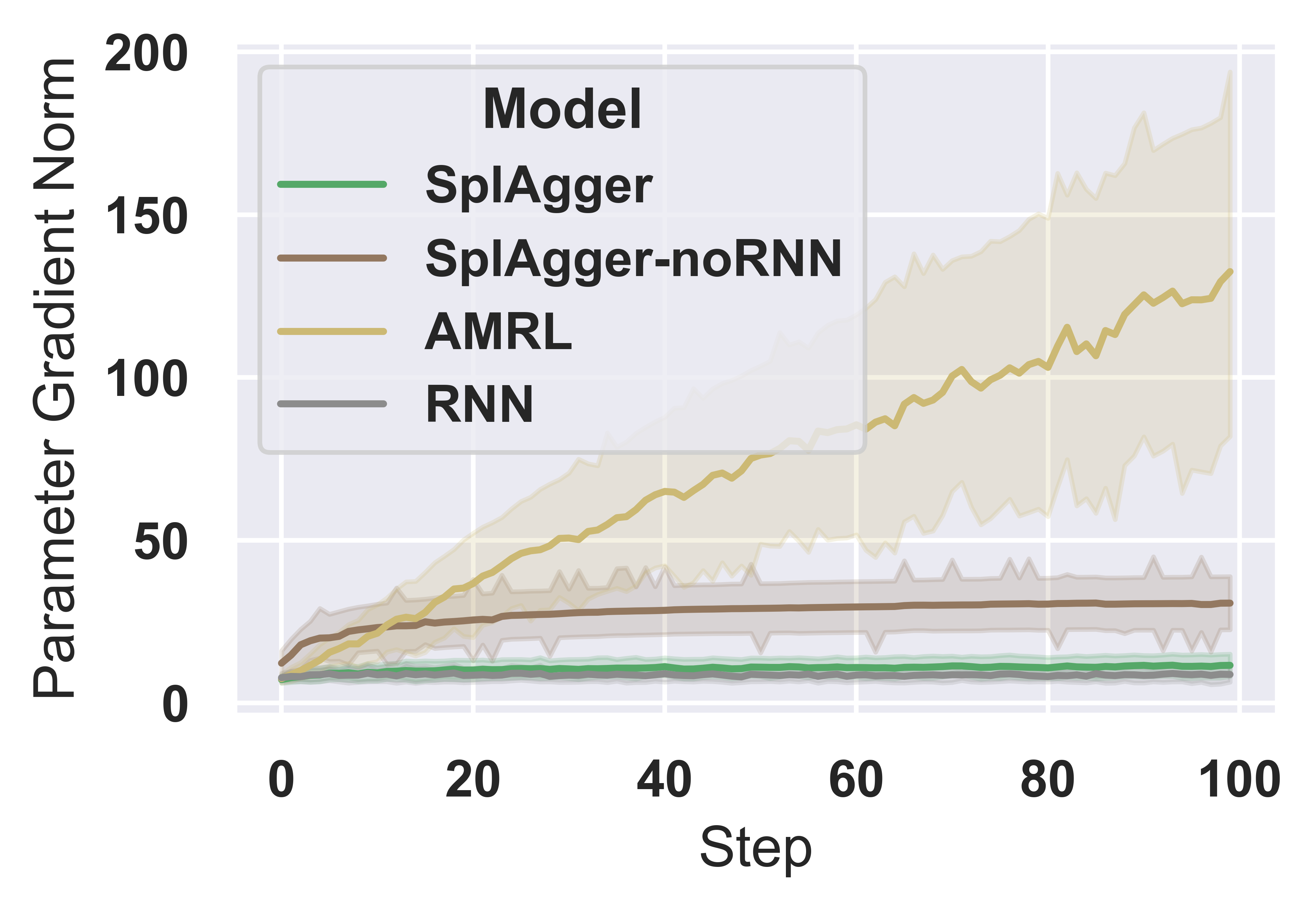}
        \caption{Gradient w.r.t. Parameters}
        \label{fig:amrl_explode}
    \end{subfigure}
    \hfill
    \begin{subfigure}[b]{0.45\textwidth}
        \centering
        \includegraphics[width=\textwidth]{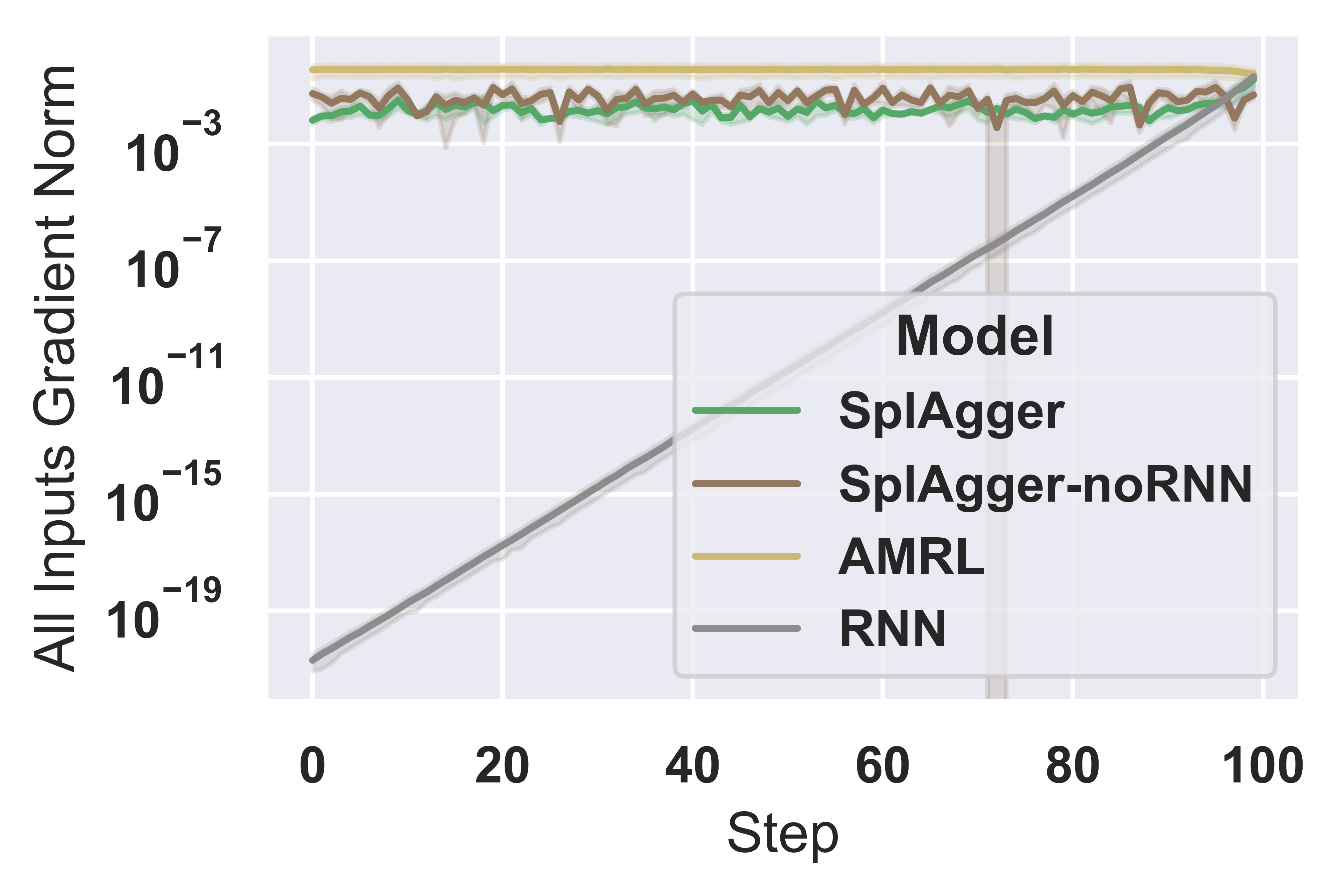}
        \caption{Final Gradient w.r.t. All Transitions}
        \label{fig:better_grad}
    \end{subfigure}
    \hfill
    \begin{subfigure}[b]{0.45\textwidth}
        \centering
        \includegraphics[width=\textwidth]{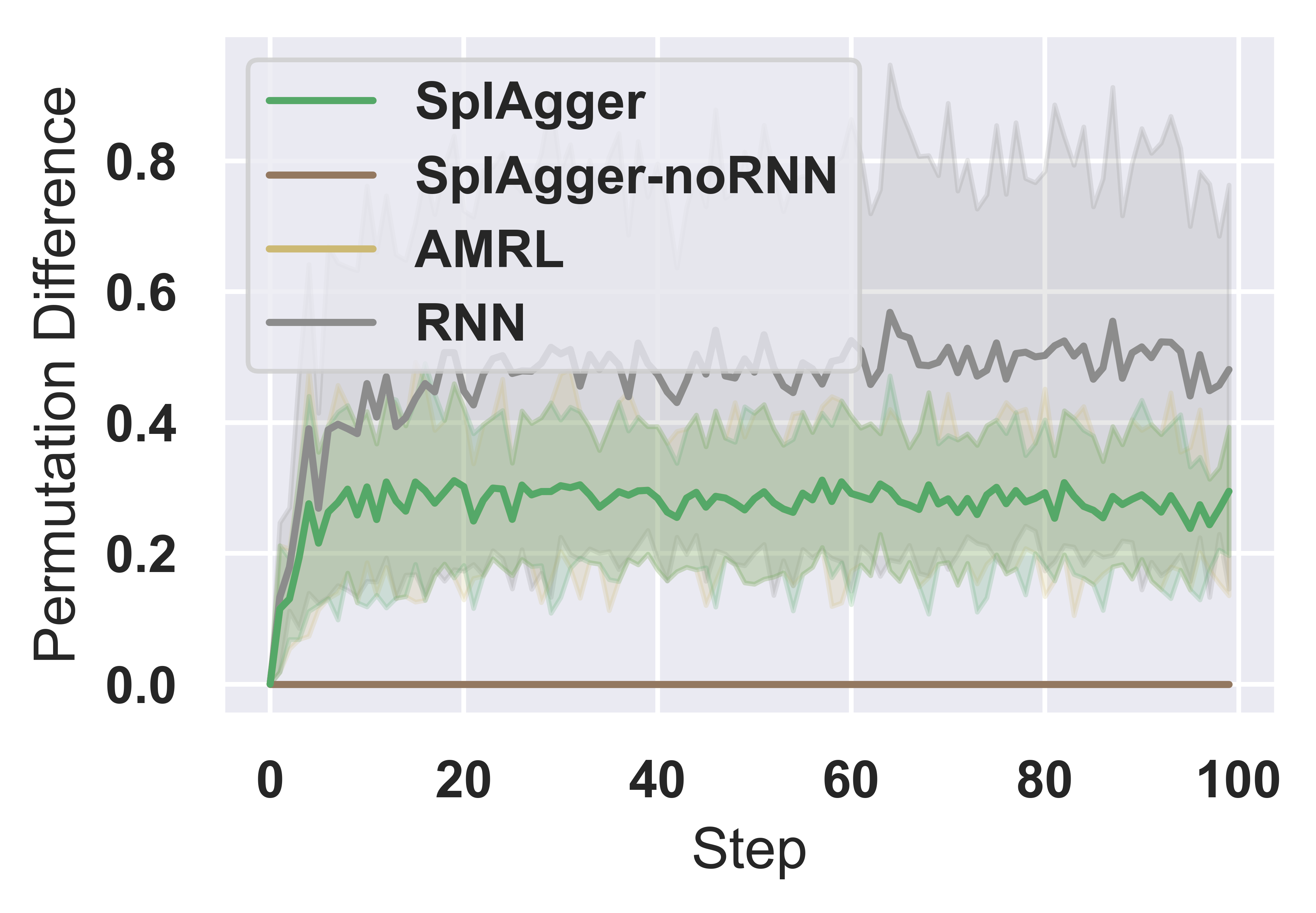}
        \caption{Encoding Permutation Difference}
        \label{fig:perm_metric}
    \end{subfigure}
    \caption{An empirical analysis of gradients in \method, AMRL, and RNNs. In \ref{fig:amrl_grad} we see that the gradient of the output of the sequence model with respect to the initial input decreases over time. AMRL modifies the gradients to prevent this, but at the cost of exploding gradients with respect to the parameters, depicted in \ref{fig:amrl_explode}. We find that poor performance of the RNN is rather due to earlier inputs having a smaller gradient, when considering the final output of the sequence model. depicted in \ref{fig:better_grad}. Finally, models that are the least permutation variant do not necessarily perform better; the highest performing model, \method, has an intermediate difference, but has components that are both permutation variant and permutation invariant. This is shown in \ref{fig:perm_metric}.
    }
    \label{fig:grads}
\end{figure}

Finally, we investigate the gradients of our sequence models to explain why, in some domains, \method works, while AMRL \citep{beck2020AMRL} does not.
AMRL demonstrates the benefit of its gradient modification by measuring the average gradient of the memory with respect to the encoding of the initial transition, $||\frac{df_t}{d\tau_0}||_2$.
AMRL shows that this quantity decays over time for normal sequence models, but not for AMRL, due to the gradient modification.
AMRL methods overwrite this gradient to set it equal to the identity.\footnote{\citet{beck2020AMRL} also note that the signal-to-noise ratio can also affect performance. However we can only recreate this result by setting the bias in all models to zero, and find it less predictive than the gradients regardless.}
We depict these gradients at initialization in Figure \ref{fig:amrl_grad}, evaluating over three model initializations.

While AMRL prevents gradient decay, it also causes gradient explosion, with respect to the model parameters, $||\frac{df_t}{d\theta}||_2$.
Since the number of inputs grows over time, and the norm of the gradients for each input does not shrink, the gradients with respect to the parameters grows.
We depict this phenomenon in Figure \ref{fig:amrl_explode}.
For an input, we sample noise uniformly over $[-1,1]$, and replicate this sample for every dimension in the input to the sequence model.

From these two gradients, it is not clear why \method performs better than an RNN, so we propose two alternative metrics for evaluation.
In Figure \ref{fig:better_grad}, we plot the gradient of the inputs, over time ($t$), holding the output time (T) fixed: $||\frac{df_T}{d\tau_t}||_2$.
This value is roughly constant for all models, except for the RNN.
For an RNN, it grows as $t$ approaches $T$, implying that, for a fixed output, the inputs become less sensitive backward in time.
In other words, by privileging recent transitions, the gradients are not permutation invariant.
The gradients are not equal for all inputs, for a given output.
In addition to this metric, we can measure the permutation variance directly.
In Figure \ref{fig:perm_metric}, we compute the mean difference between encodings of different permutations of inputs at initialization.
We also normalize the encodings, to have unit magnitude first.
We see that the RNN is the most permutation variant, and sequence models without any RNN, such as \method without an RNN, are the least.
Critically, models like \method that perform best are not the most or least permutation variant, but rather have some components of each.

\section{Conclusion}
In this paper we have shown how permutation invariance can be critical when learning to reinforcement learn.
We have, for the first time, confirmed this advantage even without the use of task inference objectives.
Surprisingly, we also demonstrate that permutation variance can still be useful, both to learn sub-optimal non-Markovian policies early on, and to make the sequence model more robust to the choice of specific aggregation function.
Using these insights, we presented \method, making use of split aggregation to achieve the best of both methods.
Moreover, we have shown that in several domains, popular existing methods fail, and discussed reasons for the failure of each.
We analyzed how the gradient modification in AMRL causes gradients with respect to the parameters to explode, and measure different types of gradient decay and permutation variance in RNNs.


\bibliography{main}
\bibliographystyle{rlc}

\appendix




\section*{Appendix}

\section{Posterior Factorization}
\label{sec:proof}
Below we include a proof of the factorization of the task posterior, as claimed in the main body.
\begin{align}
    & P(\mathcal{M}|\tau) & \\
    &= \frac{P(\tau|\mathcal{M})P(\mathcal{M})}{P(\tau)} & &\text{// Bayes's Rule} \\
    &\propto P(\tau|\mathcal{M})P(\mathcal{M}) & \\
    &= P(\tau_1,\tau_2,...,\tau_T|\mathcal{M})P(\mathcal{M}) & \\
    &= P(\mathcal{M}) \prod_{t=1}^{t=T}P(\tau_t|\tau_1,\tau_2,...,\tau_{t-1},\mathcal{M}) & \\
    &= P(\mathcal{M}) \prod_{t=1}^{t=T}P(\tau_t|s_t,\mathcal{M}) & &\text{// Markov Property} \\
    &= P(\mathcal{M}) \prod_{t=1}^{t=T}P(s_t,a_t,r_t,s_{t+1}|s_t,\mathcal{M}) & \\
    &= P(\mathcal{M}) \prod_{t=1}^{t=T}P(a_t,r_t,s_{t+1}|s_t,\mathcal{M}) &
\end{align}

\section{Hyperparameter Tuning}
\label{sec:tuning}

\begin{figure}[h]
    \centering
    \begin{subfigure}[b]{0.32\textwidth}
        \centering
        \includegraphics[width=\textwidth]{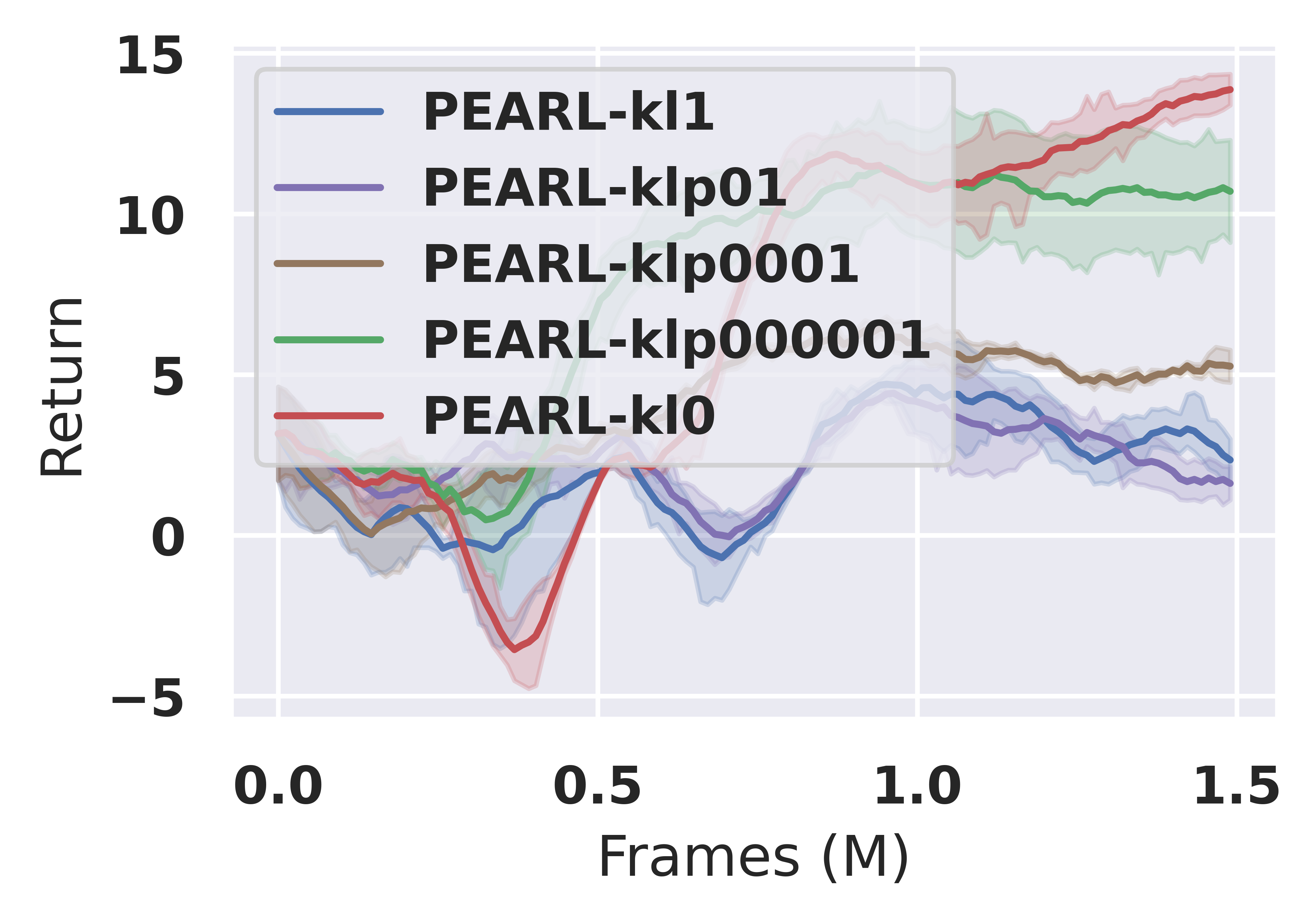}
        \caption{T-LS}
        \label{fig:pearl_tune1}
    \end{subfigure}
    \hfill
    \begin{subfigure}[b]{0.32\textwidth}
        \centering
        \includegraphics[width=\textwidth]{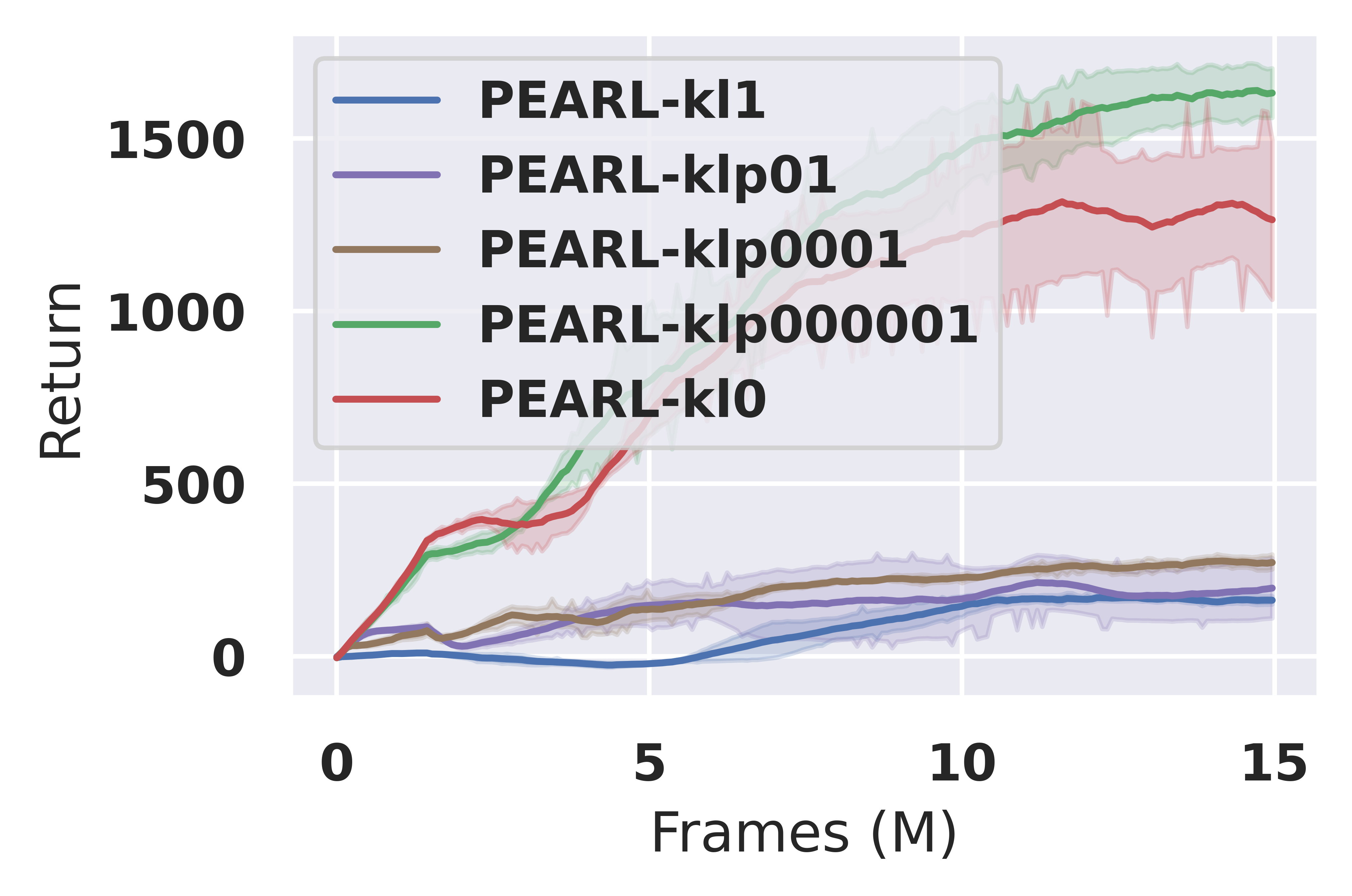}
        \caption{Walker}
        \label{fig:pearl_tune2}
    \end{subfigure}
     \hfill
    \begin{subfigure}[b]{0.32\textwidth}
        \centering
        \includegraphics[width=\textwidth]{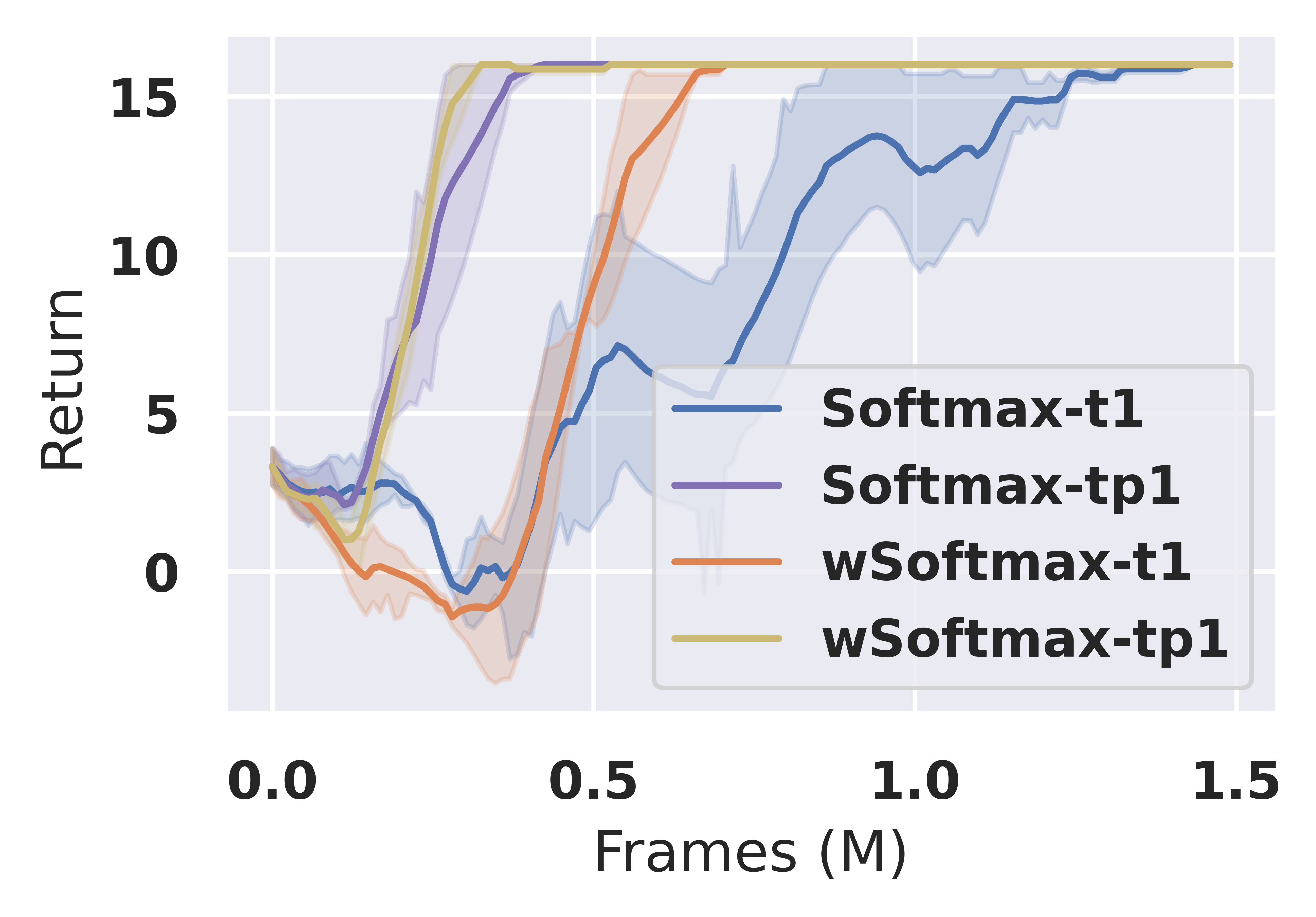}
        \caption{T-LS}
        \label{fig:temp_tune}
    \end{subfigure}
    \caption{Tuning the KL-divergence weight on PEARL, in \ref{fig:pearl_tune1} and \ref{fig:pearl_tune2}, and tuning the softmax temperatures in Figure \ref{fig:temp_tune}. For PEARL, the chosen weight, $1^{-6}$, achieved the highest returns. Using a weight of $0$ achieved a similar final returns, suggesting the stochastic latent was not particularly useful in our experiments. We chose $1^{-6}$ as it achieved the greatest average return throughout training, on both environments. For the softmax aggregators, 0.1 was the best initialization for the temperature.}
    \label{fig:pearl_tune}
\end{figure}

We tune each baseline over five learning rates for the policy, [3e-3, 1e-3, 3e-4, 1e-4, 3e-5], for three seeds.
(Results are reported with a 68\% confidence interval, computed through bootstrapping with 1,000 iterations across three seeds, consistent with all plots presented.)
Of the models evaluated, PEARL, and aggregators with the softmax aggregation, each require an additional hyperparameter. 
PEARL require a weight on objective penalizing the KL-divergence to the prior.
We tune this weight on the T-LS and Walker environments, with a weight of $1^{-6}$ performing the best.
In Figures \ref{fig:pearl_tune1} and \ref{fig:pearl_tune2}, we display these results.
Additionally, for softmax aggregators, we tuned the initial temperature of the softmax function.
The temperature is learnable, however changes very slowly, and so is sensitive to initialization.
This temperature was tuned between 1.0 and 0.1 on T-LS, with 0.1 performing the best.
Note that we tuned the aggregators without using an RNN or \method.
This is depicted in Figure \ref{fig:temp_tune}.

For all other hyperparameters, we default to those in \citet{beck2023recurrent}, with two exceptions.
First, on the Planning Game, we it necessary to set the exploration bonus in the objective to zero in order to learn any systematic exploration.
Second, \citet{beck2023recurrent} projected the output of the RNN down to size 10 or 32, and then to size 10 or 25, depending on the environment, using a linear layer, before being passed to the hypernetwork. 
For consistency, we leave this size as 24 and 25 for all models and all experiments.

\section{PEARL Design Choices and Additional Experiments}
\label{sec:pearl_additional}
\begin{figure}[h]
    \centering
    \begin{subfigure}[b]{0.32\textwidth}
        \centering
        \includegraphics[width=\textwidth]{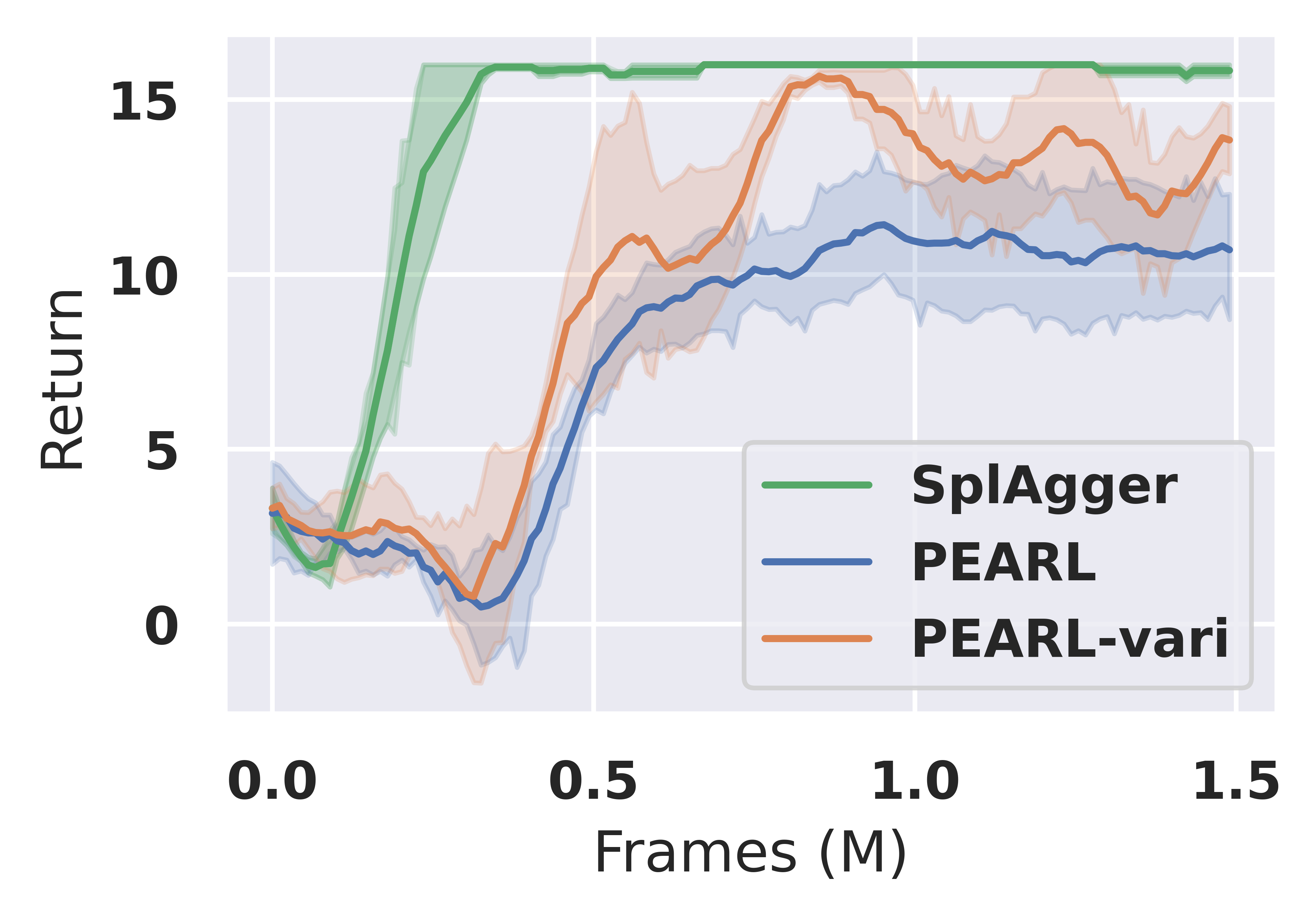}
        \caption{T-LS}
    \end{subfigure}
    \hfill
    \begin{subfigure}[b]{0.32\textwidth}
        \centering
        \includegraphics[width=\textwidth]{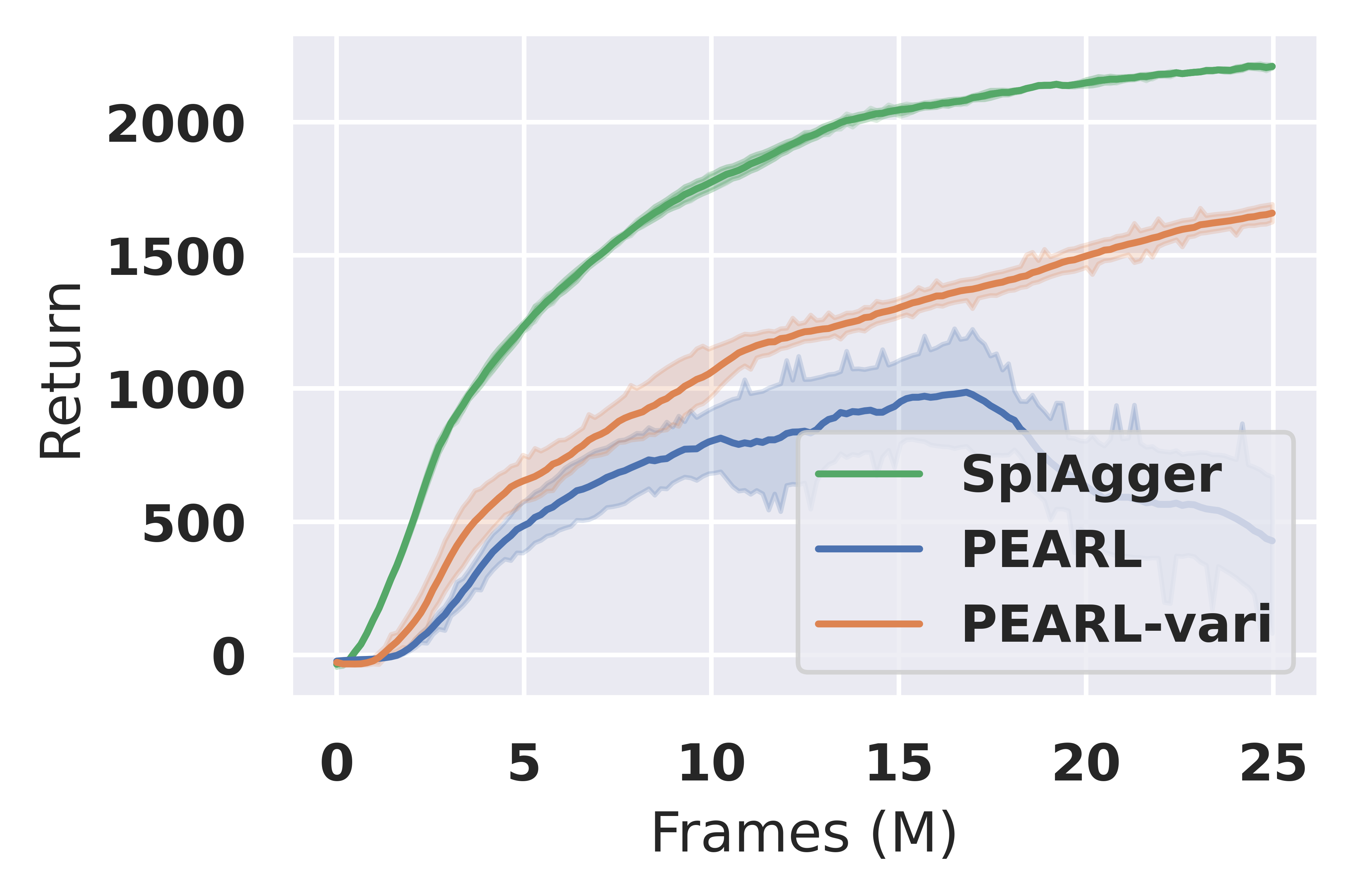}
        \caption{Cheetah-Dir}
    \end{subfigure}
    \hfill
    \begin{subfigure}[b]{0.32\textwidth}
        \centering
        \includegraphics[width=\textwidth]{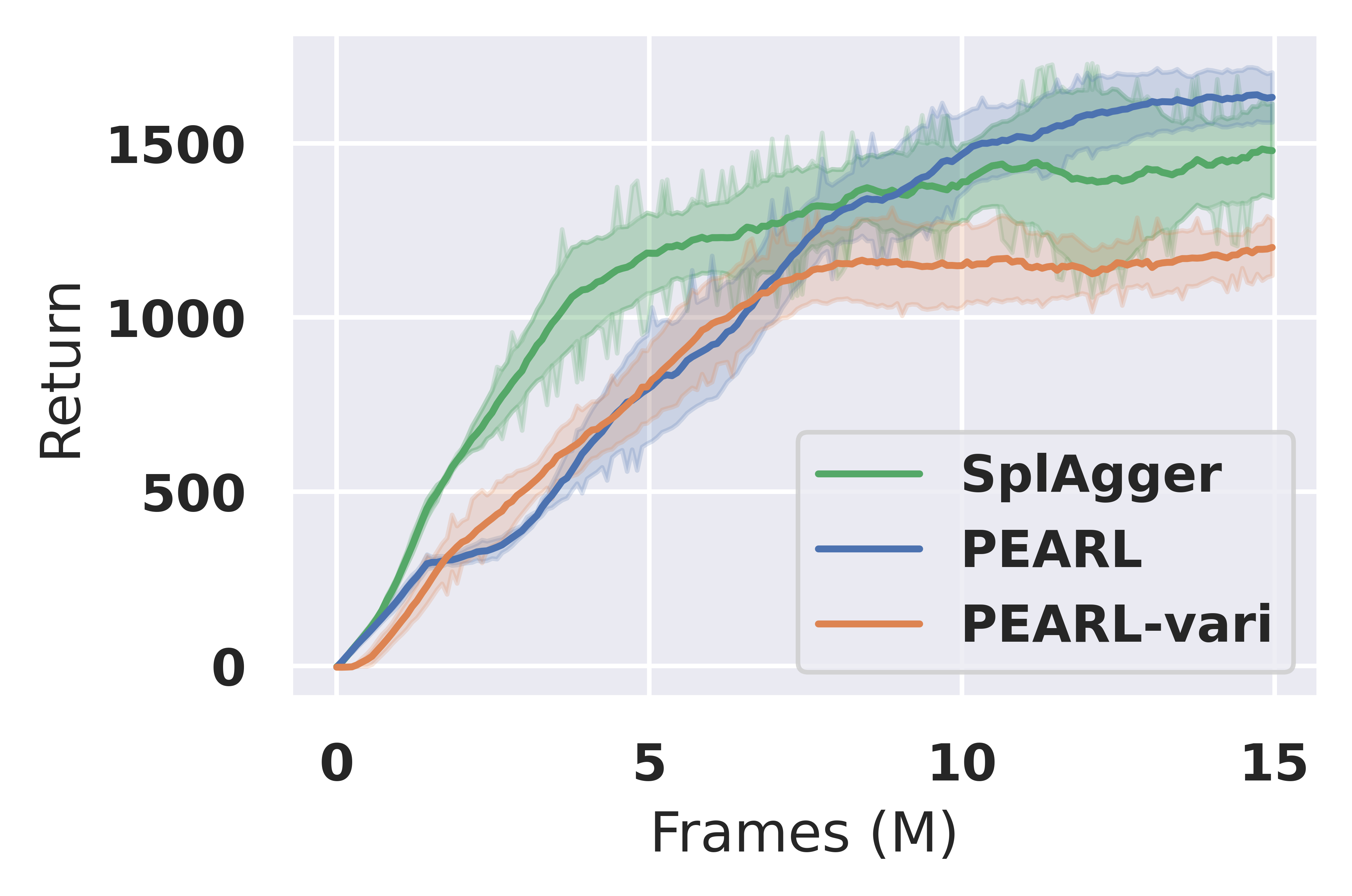}
        \caption{Walker}
    \end{subfigure}
    \caption{PEARL using the VariBad-stlye supervized reconstruction from \citet{zintgraf2021varibad} to train the sequence model (PEARL-vari). PEARL-vari, increases returns slightly on T-LS and Cheetah-Dir, but remains lower than \method.}
    \label{fig:pearl_more}
\end{figure}
In addition to the experiments with PEARL in the main body, we present a preliminary findings here.
Out of all baselines, PEARL is the only method that requires a stochastic latent variable.
In the original paper, the stochastic latent variable is sampled once per episode.
However, this type of exploration requires multiple episodes for exploration, and will necessarily fail on multiple of our benchmarks.
Instead, we consider two alternatives.
In the experiments presented in the main body of the paper, we sample the stochastic latent at every state, and here, and here we additionally present experiments that use the self-supervision provided by reconstructing rewards and transitions to train the stochastic latent, as in \citet{zintgraf2021varibad}.
We use the default hyperparameters presented by \citet{zintgraf2021varibad}.
While the results here are incomplete, and the model is still far inferior to \method, we did find some improvement, as shown in Figure \ref{fig:pearl_more}.

\section{Analysis of PEARL Aggregation Failure}
\label{sec:pearl_analysis}

\begin{figure}[h]
    \centering
    \begin{subfigure}[b]{0.42\textwidth}
        \centering
        \includegraphics[width=1\textwidth]{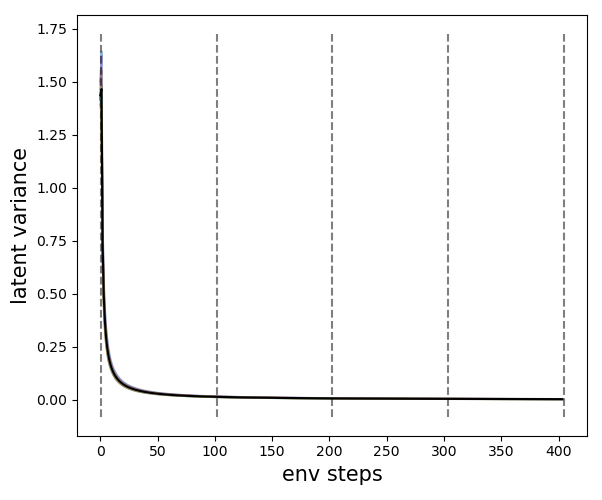}
        \caption{PEARL Latent Variance \newline (Initialization)}
        \label{fig:pearl_latent}
    \end{subfigure}
    \hfill
    \begin{subfigure}[b]{0.42\textwidth}
        \centering
        \includegraphics[width=1\textwidth]{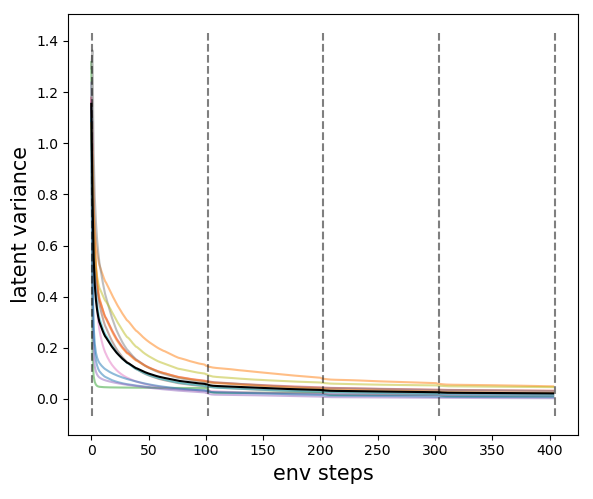}
        \caption{PEARL Latent Variance \newline (Trained)}
        \label{fig:pearl_latent_trained}
    \end{subfigure}
    \caption{In \ref{fig:pearl_latent}, we see that the variance of PEARL decreases rapidly to zero at initialization. In \ref{fig:pearl_latent_trained}, we see the same phenomenon, but when PEARL is fully trained.
    }
    \label{fig:analysis1}
\end{figure}

Here, we investigate why PEARL's aggregation performs worse than \method, and hypothesize that the poor performance has due to with modelling assumptions regarding the variance of its latent variable.
Unlike other methods, PEARL's aggregation method requires the use of a stochastic latent variable.
We observe that the variance of the aggregator decreases rapidly over time.
In Figure \ref{fig:pearl_latent} and Figure \ref{fig:pearl_latent_trained}, we can see that both at initialization and after training, the variance decreases rapidly to zero.
Looking at the probability density function (PDF), we can see why.

Specifically, PEARL uses a product of Gaussians, 
$$q_\theta(z|\tau) \propto \prod_{t = 1}^{t=T} \mathcal{N}(z;\mu_t(\tau_t),diag(\sigma^2_t(\tau_t))),$$
which has a closed form for the joint mean and covariance.
From Bayesian conjugate analysis, we know that the product of Gaussian PDFs is itself a Gaussian \citep{murphy2007conjugate}. 
The new Gaussian, $q_\theta(z|\tau)=\mathcal{N}(z;\mu',diag(\sigma'))$, has a new mean, $\mu'$, which itself is a weighted average of each individual $\mu_t$.
The respective weights are $1/\sigma^2_t$, and the variance is $\frac{1}{1/\sigma^2_0 + ... + 1/\sigma^2_T}$.
Since the denominator in the variance of PEARL grows monotonically, the variance must drop monotonically.
In fact, assuming the variances are approximately equivalent at initialization, then the variance should decrease like $1/t$.

The fact that the variances decreases over time could create problems for learning.
For example, if the agent were to collect contradictory evidence about the latent, then the variance of the true posterior would increase.
Well our domains do not present contradictory evidence, many transitions in our domains are uninformative, and so the sequence model would still need to learn to counteract the decreasing variance, which may hinder learning.
Future work could experiment with removing the latent variable from PEARL, which would be equivalent to the weighted average aggregator presented in Section \ref{sec:wavg}, or producing the variance of the latent variable using a different permutation invariant function.
Note that this pathology only applies to the aggregation method used in PEARL, and may not be the most important factor to consider in the off-policy setting in which the PEARL method originally evaluates \citep{rakelly2019efficient}.

\section{LSTM Invariant Initialization}
\label{sec:invinit}
\begin{figure}[h]
    \centering
    \begin{subfigure}[b]{0.45\textwidth}
        \centering
        \includegraphics[width=\textwidth]{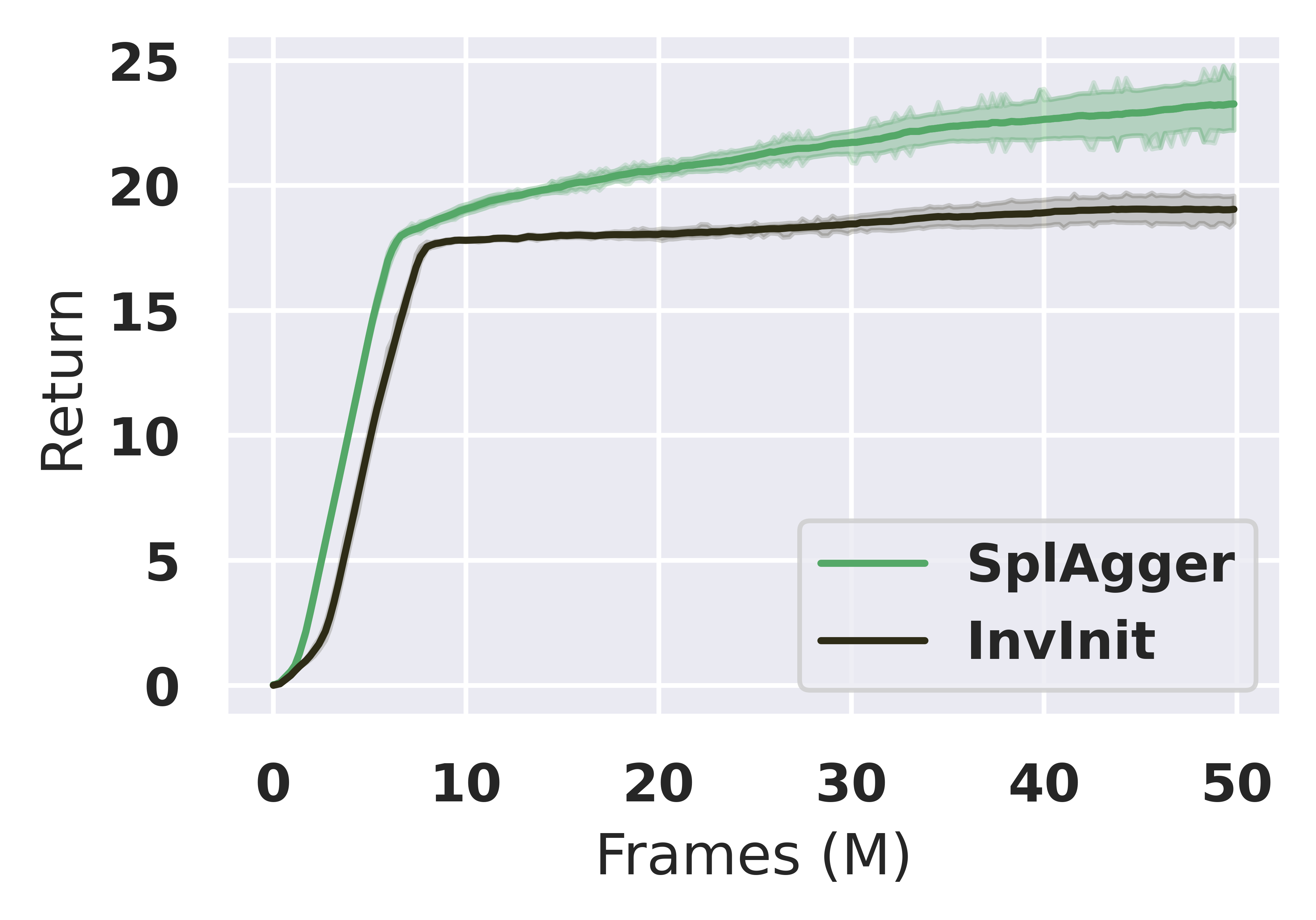}
        \caption{Planning Game}
    \end{subfigure}
    \hfill
    \begin{subfigure}[b]{0.45\textwidth}
        \centering
        \includegraphics[width=\textwidth]{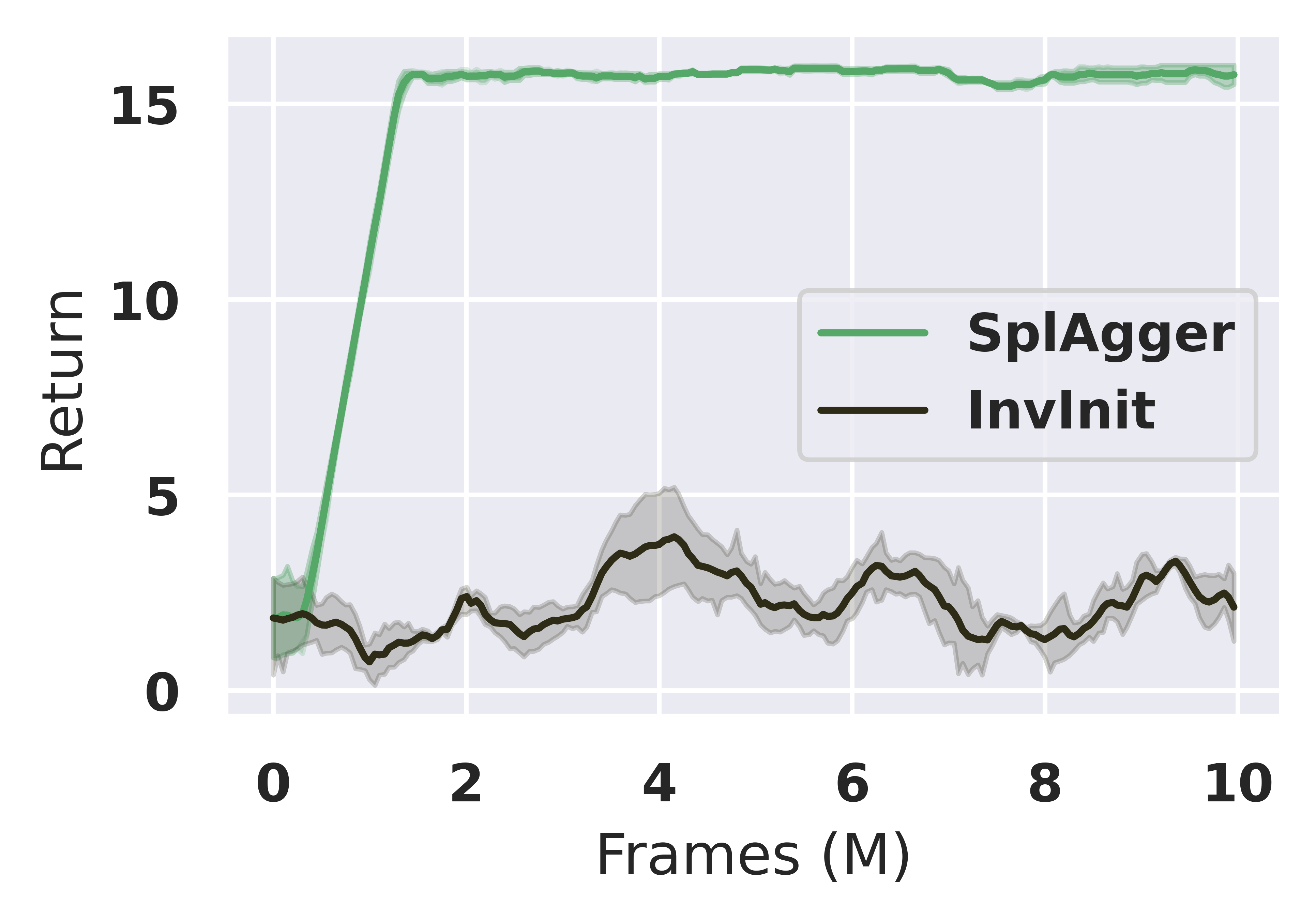}
        \caption{T-Maze Agreement}
    \end{subfigure}
    \caption{Here we evaluate an LSTM model initialized to be permutation invariant. The returns are far lower than \method.}
    \label{fig:inv_init}
\end{figure}
Finally, we experiment with using a sequence model that is initialized to be permutation invariant, but that can easily learn to be permutation variant.
In order to do this, we use a long short-term memory unit (LSTM) \citep{hochreiter1997long}, and adjust the initialization of the gates.
Specifically, we adjust the gates to compute a summation, before being normalized for the output.
First, we set the weights of the input and forget gates to zero.
Then, we adjust the bias of the input and forget gates to gate outputs of $1-\epsilon$, for $\epsilon=0.0001$.
This forces all inputs to be fully added to the running sum and not forgotten.
We additionally we set any weight connected emanating from the recurrent state, in the cell and output gates, to zero.
This forces there to be do dependence on past states.
We use an LSTM instead of a GRU so that the input gate is not forced to be the compliment of the forget gate.
Results were not encouraging, and are depicted in Figure \ref{fig:inv_init}.

\section{Transformer Results}
\label{sec:transform}
\begin{figure}[h]
    \begin{center}
        \includegraphics[width=0.5\textwidth]{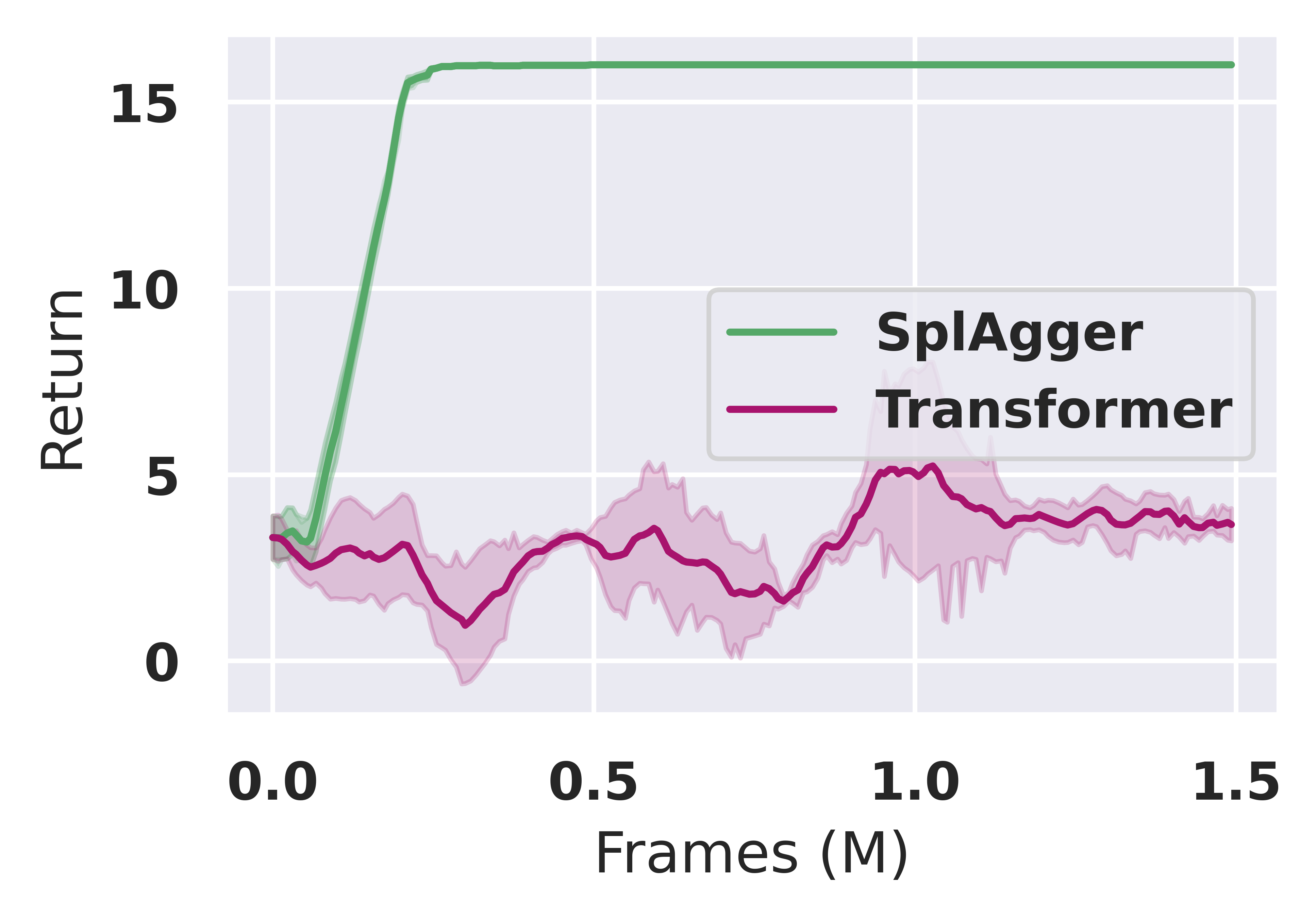}
    \end{center}
    \caption{The transformer model fails to learn and is significantly outperformed by \method on the T-LS domain.}
    \label{fig:transformer}
\end{figure}
As discussed in the main body, attention is inherently permutation invariant.
Thus, transformers without positional encodings are an obvious fit for this problem setting.
However,
Attention is computationally expensive: whereas both commutative aggregation and recurrent networks use $O(1)$ memory and compute per timestep, attention generally requires $O(t^2)$ memory and compute per timestep $t$.
We therefore consider it appropriate to limit our solutions to constant memory and compute, in line with sequence models designed to quickly handle long contexts \citep{garnelo2018neural,beck2020AMRL}.
Still, the runtime was not entirely prohibitive on one experiment: T-LS.
Thus, we did run a Transformer model on T-LS, with results depicted in Figure \ref{fig:transformer}.
The results show that the transformer did not learn in the allotted number of frames, and was significantly outperformed by \method.

\section{Additional Weighted Average Aggregation Results}
\label{sec:wavg}
\begin{figure}[h]
    \begin{center}
        \includegraphics[width=0.5\textwidth]{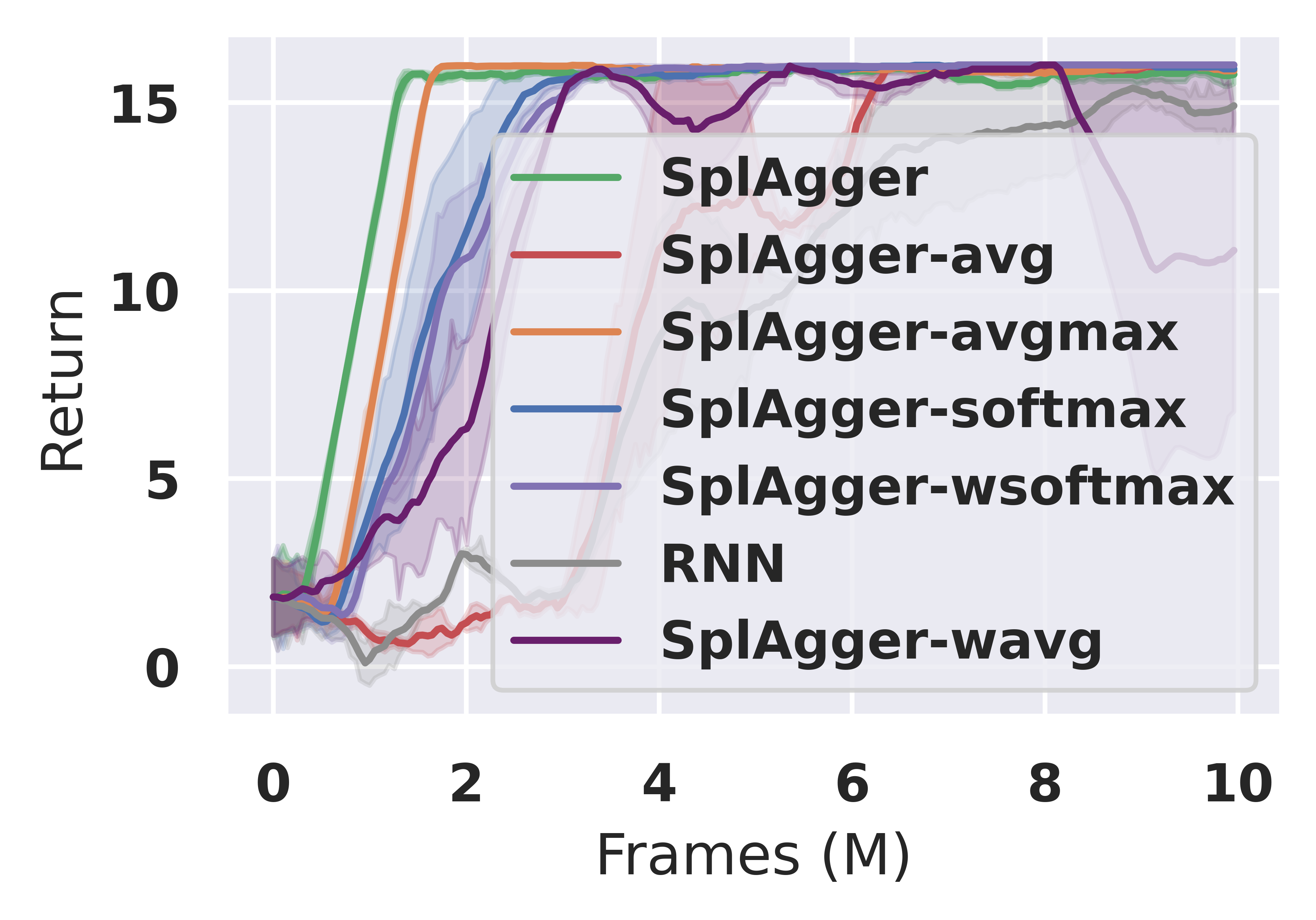}
    \end{center}
    \caption{The weighted average aggregator (wavg) performs worse than many others, including the default, max, on T-Maze Agreement.}
    \label{fig:wavg}
\end{figure}
In addition to the softmax aggregator, which computes a weighted softmax of the inputs, we experimented with a weighted average aggregator that does not use softmax.
The weights and aggregates are still predicted separately for each input, as with wsoftmax aggregator, resulting in half the number of neurons output as are input.
We chose to evaluate this aggregator as it more closely resembles the aggregator in PEARL.
In fact, it is the same aggregation method if PEARL always output zero variance, and thus did not train a stochastic latent variable.
Our results, shown in figure \ref{fig:wavg} show this aggregation method to learn slightly faster than the average aggregator, but less stable.
It also performed worse than max, softmax, and wsoftmax.

\section{Environment Details}
\label{sec:env_details}

\paragraph{Cheetah-Dir.}
In Cheetah-Dir \cite{zintgraf2021varibad}, the agent outputs six different torques in order to control a cheetah robot.
The agent is rewarded for its velocity in either the forward or reverse direction, depending on the sampled task, and receives a penalty for the magnitude of the torques.
The agent receives a 17-dimensional observation consisting of the position, angle, and velocity of each body part.
Here, the agent has one episode during which it can adapt to the task.

\paragraph{Walker.}
In Walker \cite{zintgraf2021varibad}, the agent outputs six different torques in order to control a two-legged torso morphology.
The agent receives 17-dimensional observations and receives a reward for running only in the forward direction.
The tasks are defined by random samples of 65 different physics coefficients, such as body mass and friction, which collectively define the task.
Here, the agent has two episodes during which it can adapt to the task.

\paragraph{T-LS.}
In the T-LS environment \cite{beck2020AMRL}, the agent inhabits a T-shaped maze, called a T-Maze.
The agent is shown a signal and then deterministically stepped down a corridor of length 100.
At the end of the corridor, that agent opens one of two doors.
If the door matches the signal, then the agent receives a reward of 4, and if not, it receives a penalty of -3.
In between the start and end of the corridor, observations are augmented by a Bernoulli random variable. 
The agent adapts across four sequential episodes, which together constitute a single meta-episode.

\paragraph{MC-LS.}
The MC-LS environment \cite{beck2020AMRL} is designed to challenge an agent's long-term memory based on visual cues from Minecraft. The task involves navigating a sequence of 16 rooms. In each room, the agent must choose to go either left or right around a central column. This decision is based on the column's material: diamond or iron. Discrete actions allow for a finite set of observations. When the agent makes a decision in line with the observed material, it earns a reward of 0.1. The final decision at the end of the sequence is dictated by a color signal (red or green) presented at the beginning, which specifies the task. As in the T-Maze, the agent receives a high reward (4) for a correct final action and a significant penalty (-3) for an incorrect one. The agent adapts across two sequential episodes, which together constitute a single meta-episode. 

\paragraph{Planning Game.}
The Planning Game, which is adapted from \citet{ritter2021rapid}. 
In the Planning Game \citep{ritter2021rapid}, the agent inhabits a $3\times3$ grid that wraps around each side.
In this grid, the agent must navigate to a goal.
The goal location, observed by the agent, changes multiple times during a single episode.
The MDP is not defined by the goal location, shown to the agent, but rather by a changing transition function.
The state in each cell of the grid changes in each task.
A task is thus defined by a permutation of the 9 states.
For instance, while the agent may be in the bottom right of the grid in two different tasks, it observes a different state there in each task.
The states themselves are encoded by one-hot encodings.
The agent must explore each state once to learn where each state is on the grid.
Once it has seen each state, then it can immediately interpret the given goal instruction.
Without knowing the state locations, there is a suboptimal policy that does not require remembering each state, namely, re-exploring the entire grid every time a goal is reached. 
Note the grid size in our experiments is $3\times3$, rather than the original $4\times4$, to decrease the total number of frames required for training.

\section{Additional Baselines}
\label{sec:alt_agg}

\paragraph{\method-noRNN-avg.} This method replaces the max operator in \method with an average and removes the RNN.
Since removal of the RNN obviates the need for split aggregation, split aggregation is removed as well.
This is equivalent to just computing a mean operation over linear encodings of each transition, $\tau_t$.

\paragraph{\method-avgmax.} This method replaces the max operator in \method with an average and a max operator.
The avgmax aggregator averages half of the neurons and computes a maximum over the other half.
This aggregation is evaluated as a way to use both the average and max aggregation.

\paragraph{\method-softmax.} This method replaces the max operator in \method with a softmax operator.
The softmax aggregator computes an average over the inputs where the weights are determined by the softmax of the aggregates themselves.
In order to aggregate online, in $O(1)$ memory, we store both the sum of the weighted aggregates, i.e., $n=e_0\textrm{exp}(e_0/\eta)+...+e_t\textrm{exp}(e_t/\eta)$, and the sum of the weights seen so far, i.e., $d=\textrm{exp}(e_0/\eta)+...+\textrm{exp}(e_t/\eta)$, where $\eta$ is a learnable temperature parameter.
The output is then the quotient of the weighted aggregates and the weights, $n/d$.
This aggregator interpolates between the average and max.
The initialization of the temperature is important, as it defines the interpolation, and is set to 0.1.
Tuning information is available in Appendix \ref{sec:tuning}.

\paragraph{\method-wsoftmax.} This method replaces the max operator in \method with a different softmax operator.
This aggregator is the same as softmax, but the weights are predicted separately from the weighted encodings.
Half of the input neurons are aggregated, and the other half are used to compute the logits for the softmax weights.
An additional version of this aggregator without the softmax is investigated in Appendix \ref{sec:wavg}.

\end{document}